%% file: main.tex
\pdfoutput=1

\documentclass[11pt]{article}

\usepackage[preprint]{acl}

\usepackage{times}
\usepackage{latexsym}
\usepackage{graphicx}
\usepackage{enumitem}
\usepackage{cuted}
\usepackage{dblfloatfix}

\usepackage[T1]{fontenc}

\usepackage[utf8]{inputenc}

\usepackage{microtype}

\input{preamble}

\usepackage{subfigure}
\usepackage{nicefrac}
\usepackage{float}
\usepackage{times}
\usepackage{latexsym}
\usepackage{amsmath}
\usepackage{amsfonts}
\usepackage{amssymb}
\usepackage{algorithm}
\usepackage{algpseudocode}
\usepackage{algorithm}
\usepackage{algpseudocode}
\usepackage{graphicx}
\usepackage[most]{tcolorbox}
\usepackage{fontawesome5} 
\usepackage{caption}      

\usepackage{enumitem}
\setlist[itemize]{noitemsep, topsep=0pt, leftmargin=*}

\newlist{compactitem}{itemize}{1}
\setlist[compactitem]{noitemsep, topsep=0pt}

\newlist{compactenum}{enumerate}{1}
\setlist[compactenum]{noitemsep, topsep=0pt, label=\arabic*.}

\usepackage{tcolorbox}

\newtcolorbox{promptbox}{
    enhanced,
    breakable,
    colback=cyan!4,
    colframe=cyan!40!black,
    boxrule=0.4pt,
    arc=2pt,
    left=4pt,
    right=4pt,
    top=4pt,
    bottom=4pt,
    width=\linewidth
}

\DeclareMathAlphabet{\mathbcal}{OMS}{cmsy}{b}{n}
\usepackage{amsmath}
\usepackage{mathrsfs}
\usepackage{comment}

\usepackage{bm}
\usepackage{bbm}
\usepackage{amssymb}

\newcommand{\nish}[1]{\textcolor{magenta}{Nish:~#1}}

\title{Not Worth Another Token:\\
Marginal Value Estimation 
for Efficient Deep Research Agents\vspace{0.5em}}

\author{
Harshitha Kolukuluru$^{1}$, Reshma Ashok$^{1}$, Kirat Arora$^{1}$\\
{\bf Evan William Ciccarelli$^{1}$}, {\bf Nischal Ashok Kumar$^{1}$}, {\bf Lunyiu Nie$^{2}$}\\
{\bf Franck Dernoncourt$^{3}$}, {\bf Samyadeep Basu$^{3}$}, {\bf Ryan A. Rossi$^{3}$}, {\bf Nedim Lipka$^{3}$}\\[0.5em]
$^{1}$University of Massachusetts Amherst\\
$^{2}$The University of Texas at Austin\\
$^{3}$Adobe Research\\
\texttt{\{hkolukuluru,rashok\}@umass.edu, lynie@utexas.edu, lipka@adobe.com}
}
\begin{document}
\maketitle
\begin{abstract}
Long-horizon research agents solve open-ended tasks through iterative retrieval, aggregation, and synthesis, but context grows rapidly while the marginal value of additional evidence often declines. This leads to unnecessary token costs, higher latency, and noisier inputs for final report generation. We study \emph{marginal value estimation} for context management in a GPT-Researcher-style deep research pipeline, comparing pruning strategies at pre-retrieval, post-retrieval, and pre-synthesis stages. Across lightweight heuristic criteria, LLM-based pruning, and an exploratory learned controller, we find that pruning effectiveness depends strongly on \emph{where} pruning is applied: early pruning yields the largest end-to-end savings, while later pruning mainly refines the final synthesis context. Lightweight heuristics reduce token usage by up to 73\% while approximately preserving judged report quality under a fixed evaluation setup. No single method dominates across efficiency, judged quality, relevance, and source-grounding proxies. These findings provide practical guidance for designing more efficient long-horizon research agents.
\end{abstract}

\input{sections/01_introduction}

\input{sections/02_related_work}

\input{sections/03_method}
\input{sections/04_information_gain}

\input{sections/06_experiments}

\input{sections/07_conclusion}
\bibliography{main}
\bibliographystyle{acl_natbib}

\appendix

\input{sections/05_metric_computation}

\input{sections/09_prompts}

\input{sections/10_others}

\input{sections/11_graphs}
\end{document}

%% file: preamble.tex
\usepackage{xcolor}
\definecolor{thedarkblue}{RGB}{0,0,120} 
\definecolor{mydarkblue}{rgb}{0,0.08,0.45} 
\definecolor{darkblue}{rgb}{0,0.08,180}
\colorlet{TufteRed}{red!80!black}
\definecolor{theblue}{RGB}{0,0,180}
\colorlet{thered}{TufteRed}
      
\usepackage{microtype}
\usepackage{balance}

\usepackage{booktabs}
\usepackage{tabularx}

\usepackage{amsmath,amssymb,amsthm}

\newcommand{\eat}[1]{\ignorespaces}
\usepackage{comment}

\newcommand{\journal}[1]{} 

\usepackage{tikz}
\usepackage{verbatim}
\usetikzlibrary{arrows}
\usetikzlibrary{shapes,snakes}
\usetikzlibrary{decorations.pathmorphing} 
\usetikzlibrary{fit}					
\usetikzlibrary{backgrounds}	

\usepackage{ragged2e}
\usepackage{multirow}
\usepackage{microtype}
\usepackage{balance}
\usepackage{setspace}

\graphicspath{{./}{./graphics/}}
\newcolumntype{H}{>{\setbox0=\hbox\bgroup}c<{\egroup}@{}}

\newcolumntype{R}[1]{>{\RaggedLeft\arraybackslash}} 
\newcolumntype{L}[1]{>{\RaggedRight\arraybackslash}} 

\AtBeginEnvironment{pmatrix}{\setlength{\arraycolsep}{2pt}}

\DeclareMathOperator{\hugeE}{\mbox{\huge\raise-0.3ex\hbox{E}}}
\DeclareMathOperator{\p}{\mathbb{P}}
\DeclareMathOperator{\hugep}{\mbox{\huge\raise-0.3ex\hbox{$\p$}}}



%% file: sections/01_introduction.tex
\section{Introduction}
\label{sec:intro}

Long-horizon retrieval agents answer complex, open-ended queries through iterative retrieval, reasoning, and synthesis \cite{yao2023react,yao2023tree,nie2025flashresearch}. As these systems operate over multiple steps, accumulated context grows rapidly while the marginal value of each additional retrieval often declines: early steps surface core facts, whereas later ones add redundant or weakly informative content \cite{liu2024lost,jiang2024longllmlingua,xu2023recomp}, increasing token cost, latency, and noise in the final synthesis context.

A natural remedy is to prune low-value candidates by estimating their \emph{marginal value}---how much new, useful information each contributes beyond what is already collected. Existing deep-research pipelines often rely on LLM prompting for these decisions, incurring inference overhead and inconsistent behavior \cite{nie2025flashresearch}. Yet there is limited empirical understanding of \textbf{which pipeline stages benefit most from pruning, which strategies work best at each stage, and under which objectives}.

We address this through a stage-aware analysis of marginal-value-based pruning in deep research pipelines (Figure~\ref{fig:deep_research_pipeline}), where a system decomposes a query into subqueries, accumulates context across retrieval steps, and synthesizes a long-form report. We frame pruning as a stage-aware design problem: marginal-value criteria should be evaluated by both how they score candidates and where they intervene in the research workflow. We study three intervention points, \emph{Pre-Retrieval}, \emph{Post-Retrieval}, and \emph{Pre-Synthesis}, which affect different parts of search expansion, context growth, and final synthesis, and compare heuristic, learned, and LLM-based strategies across one-, two-, and three-stage configurations under a shared evaluation setup (Sections~\ref{sec:problem}-\ref{sec:methods}).


Stage placement is often more consequential than the specific scoring rule. Post-Retrieval MMR reduces token cost from 375.4k to 114.6k and explored nodes from 29.0 to 8.84 with modest quality loss; Pre-Synthesis pruning is generally too late to recover upstream costs but can improve report quality. Two-stage pruning yields the strongest quality-efficiency trade-offs; three-stage MMR achieves 73.3\% token reduction. No single strategy dominates: relevance--redundancy heuristics excel at early cost control, while richer methods are more competitive when targeting report quality or source grounding.

The main contributions of this work are:
\begin{itemize}
    \item We formulate marginal value estimation as a stage-aware pruning problem for deep research pipelines, covering three intervention points:
    \emph{Pre-Retrieval}, \emph{Post-Retrieval}, and \emph{Pre-Synthesis}.
    \item We provide a controlled empirical comparison of heuristic, learned, and LLM-based pruning strategies across one, two, and three-stage configurations under a shared evaluation setup.
    \item We identify cross-stage findings about where pruning is most effective, which strategies provide the best quality-cost trade-offs, and where report quality, source grounding, and cost reduction diverge.
\end{itemize}

%% file: sections/02_related_work.tex
\section{Related Work}
\label{sec:related_work}

Prior work on long-horizon retrieval agents studies how language models can decompose tasks, retrieve iteratively, and reason over multi-step search trajectories. Tree of Thoughts \cite{yao2023tree} and ReAct \cite{yao2023react} introduced branching and tool-augmented reasoning, while ParallelResearch \cite{nie2025flashresearch} extends this direction to deep research through tree-structured query decomposition and parallel exploration. These systems improve multi-step retrieval and coverage, but primarily focus on how to explore rather than how to control the growth of accumulated context.

A separate line of work studies context pruning, compression, and prompt-efficiency methods for long-context language models. Lost in the Middle \cite{liu2024lost} shows that models struggle to use relevant information in long prompts, motivating stronger context management. Selective Context \cite{li2023selectivecontext}, DYCP \cite{choi2026dycp}, LLMLingua and LongLLMLingua \cite{jiang2023llmlingua,jiang2024longllmlingua}, and RECOMP \cite{xu2023recomp} reduce prompt length or compress retrieved evidence, typically at or near the final model input. Related work on agent orchestration and adaptive control studies how multi-step pipelines should schedule actions, react to intermediate signals, or terminate search \cite{agentwarpp2025,nalar2026,cortex2025,shinn2023reflexion,reseek2025}. Our focus is complementary: we study marginal-value-based pruning as a stage-aware context management problem, and compare pruning strategies across \emph{Pre-Retrieval}, \emph{Post-Retrieval}, and \emph{Pre-Synthesis} under a shared evaluation setup.

\begin{figure*}[t]
    \centering
\includegraphics[width=\textwidth]{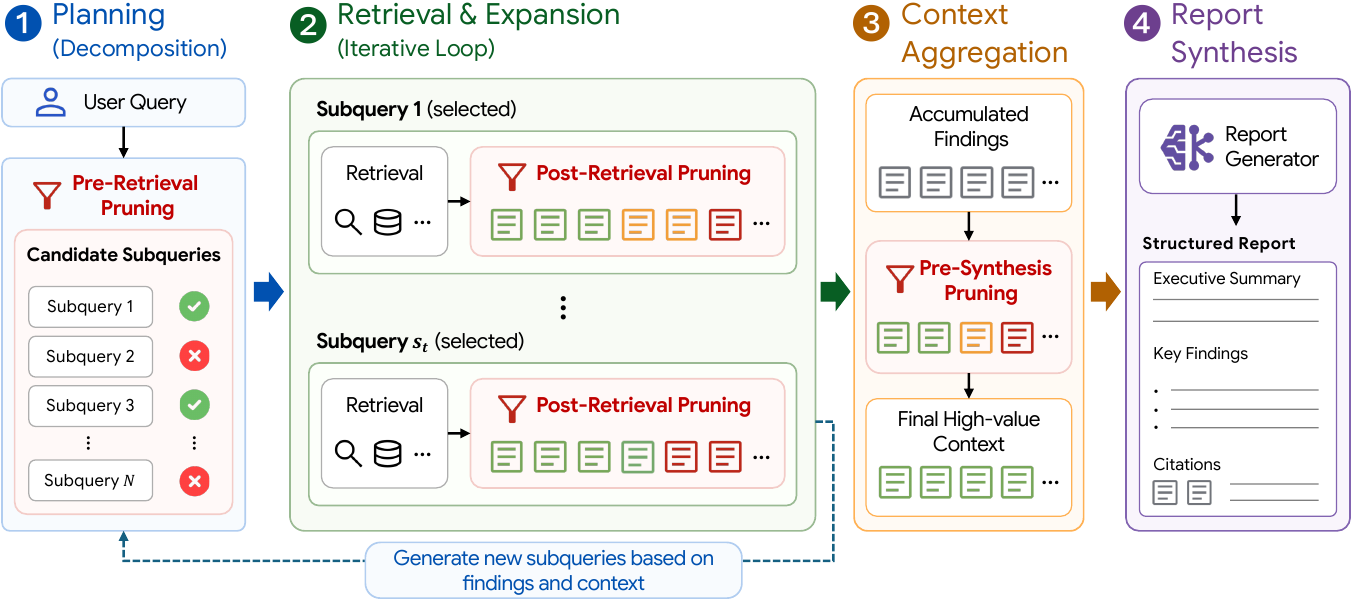}
\caption{Overview of the deep research pipeline. Given a user query, the system proceeds through four stages: (1) planning via query decomposition, (2) retrieval and branch expansion, (3) context aggregation, and (4) final report synthesis. We study pruning at three intervention points: \emph{Pre-Retrieval}, which filters candidate subqueries before search; \emph{Post-Retrieval}, which filters retrieved context during branch expansion; and \emph{Pre-Synthesis}, which compresses the accumulated context before report generation.}

    \label{fig:deep_research_pipeline}
\end{figure*}

%% file: sections/03_method.tex
\section{Problem Formulation}
\label{sec:problem}

\subsection{Deep Research Pipeline with Pruning}

We consider long-horizon retrieval agents that answer complex queries through iterative decomposition, retrieval, aggregation, and synthesis, instantiated here as a tree-structured deep research workflow~\cite{nie2025flashresearch} (Figure~\ref{fig:deep_research_pipeline}).

Given query $Q$, the system decomposes $Q$ into subqueries $\mathcal{S}$. At step $t$, it selects $s_t \in \mathcal{S}$, retrieves context items $C_{s_t}$, and merges retained evidence into accumulated context $C_t$. Retrieved findings may spawn additional subqueries, extending $\mathcal{S}$ dynamically. The process terminates when $\mathcal{S}$ is exhausted or a stopping criterion is met at step $T$, after which the accumulated context is passed to synthesis.

\noindent\textbf{Objective.} Let $\mathcal{R}(C_T, Q)$ denote report quality and $\operatorname{Cost}(C_T)$ operational cost (token usage, retrieval calls, latency). In the unpruned setting, evidence is added unconditionally:
\[
C_{t+1} = C_t \cup C_{s_t}.
\]
Since not all retrieved content contributes equally, we seek a policy that maximizes quality subject to a cost budget $B$:
\[
\max_{C_T \subseteq \bigcup_t C_{s_t}} \mathcal{R}(C_T, Q)
\quad
\text{s.t.}
\quad
\operatorname{Cost}(C_T) \le B,
\]
or equivalently, via Lagrangian relaxation,
\[
\max_{C_T} \; \mathcal{R}(C_T,Q) - \eta\,\operatorname{Cost}(C_T),
\]
where $\eta \ge 0$ selects an operating point on the quality-efficiency frontier. (We use $\eta$ to avoid overloading $\lambda$, which later denotes the MMR trade-off parameter.)

Joint optimization over all subsets is intractable, so we decompose the problem into local pruning decisions at three intervention points:

\begin{itemize}
    \item \textbf{Pre-Retrieval.} A candidate subquery $s_t \in \mathcal{S}$ is scored by predicted marginal value; low-value subqueries are filtered before incurring retrieval cost.
    \item \textbf{Post-Retrieval.} Each retrieved item $c \in C_{s_t}$ is evaluated against $C_t$; low-value items are discarded before influencing further branch expansion.
    \item \textbf{Pre-Synthesis.} The final context $C_T$ is pruned to retain only high-value evidence before report generation.
\end{itemize}

\paragraph{Stage combinations.}
A \emph{one-stage} configuration prunes at exactly one point (\emph{Post-Retrieval} or \emph{Pre-Synthesis}); a \emph{two-stage} configuration combines \emph{Post-Retrieval} + \emph{Pre-Synthesis}; a \emph{three-stage} configuration prunes at all three points. We exclude \emph{Pre-Retrieval}-only and other partial \emph{Pre-Retrieval} combinations from the one- and two-stage analyses: \emph{Pre-Retrieval} decisions are necessarily more predictive and error-sensitive than later-stage decisions conditioned on retrieved context. This decomposition lets us isolate whether gains arise from reducing search expansion, compressing the final synthesis context, or both.

\subsection{Marginal Value Estimation as a Decision Problem}

At each pruning point, the system decides whether candidate $x$ contributes sufficient marginal value to retain, given query $Q$ and accumulated context $C_t$: at \emph{Pre-Retrieval}, $x$ is a subquery $s \in \mathcal{S}$; at \emph{Post-Retrieval} and \emph{Pre-Synthesis}, $x$ is a context item $c$. We formalize this through a unified scoring function $\mathcal{V}(x \mid C_t, Q)$, retaining $x$ if its score exceeds a stage-specific threshold ($\tau_{\text{pre}}$, $\tau_{\text{post}}$, or $\tau_{\text{syn}}$). Because relevance, novelty, diversity, and coverage may matter differently at different stages, selecting the right instantiation of $\mathcal{V}$ motivates the cross-stage comparison in Section~\ref{sec:methods}. Algorithm~\ref{alg:orchestration} shows the full pipeline.
\begin{algorithm}[t]
\small
\caption{Deep Research with Context Pruning}
\label{alg:orchestration}
\begin{algorithmic}[1]

\Require Query $Q$; thresholds $\tau_{\text{pre}}, \tau_{\text{post}}, \tau_{\text{syn}}$
\State $C_0 \gets \emptyset$, $\mathcal{S} \gets \mathrm{Decompose}(Q)$, $t \gets 1$

\While{$\mathcal{S} \neq \emptyset$}
    \State Select and remove subquery $s_t \in \mathcal{S}$
    \State \(\triangleright\) \textit{Pre-retrieval pruning}
    \If{$\mathcal{V}(s_t \mid C_{t-1}, Q) < \tau_{\text{pre}}$}
        \State $C_t \gets C_{t-1}$
    \Else
        \State Retrieve context items $C_{s_t}$ using $s_t$
        \State \(\triangleright\) \textit{Post-retrieval pruning}
        \State $C_{s_t} \gets \{c \in C_{s_t} : \mathcal{V}(c \mid C_{t-1}, Q) \ge \tau_{\text{post}}\}$
        \State $C_t \gets C_{t-1} \cup C_{s_t}$
        \State Add new subqueries spawned from $C_{s_t}$ to $\mathcal{S}$
    \EndIf
    \State $t \gets t + 1$
\EndWhile

\State \(\triangleright\) \textit{Pre-synthesis pruning}
\State $\widehat{C}_T \gets \{c \in C_T : \mathcal{V}(c \mid C_T, Q) \ge \tau_{\text{syn}}\}$
\State Generate final report from $\widehat{C}_T$

\end{algorithmic}
\end{algorithm}

\section{Motivating Analysis}
\label{sec:motivating_analysis}



The unpruned pipeline is expensive by design: it explores \textbf{29.0} nodes, consumes \textbf{375.4k} tokens, and requires \textbf{3422.6s} per report (Table~\ref{tab:quality_one_stage_main}). A key source of inefficiency is that low-value context survives until late in the pipeline. The unpruned pipeline's built-in pre-synthesis trimming reduces accumulated context from \textbf{66.10} to \textbf{44.08} items (\textbf{33.3\%} fewer items; \textbf{34.06\%} fewer tokens), meaning substantial retrieved material is discarded only after most retrieval and processing cost has been paid. Since token cost is dominated by retrieval and result processing rather than planning or query generation, late trimming compresses the synthesis prompt but cannot recover earlier search costs. This motivates our central question: can low-value context be removed earlier, and does  intervening at \emph{Pre-Retrieval}, \emph{Post-Retrieval}, or \emph{Pre-Synthesis}, meaningfully affect quality and cost trade-offs?

%% file: sections/04_information_gain.tex
\section{Pruning Strategies}
\label{sec:methods}

We compare heuristic, lexical, LLM-based, and learned pruning strategies, each instantiating a different notion of marginal value under the stage-aware framework of Section~\ref{sec:problem}. 

Let \(e(\cdot)\in\mathbb{R}^d\) denote the embedding function, and let
\[
q = e(Q)
\]
be the root-query embedding. Candidate items are denoted by \(x\), and retained context by \(C\). When a formula operates in embedding space, we write \(e(x)\) for the embedding of candidate \(x\) and \(e(c)\) for the embedding of a retained item \(c\in C\). We use cosine similarity
\[
\mathrm{sim}(u,v) = \frac{u^\top v}{\|u\|\,\|v\|}
\]
for vector inputs \(u,v\). For methods that require a nonnegative scalar query-relevance weight, we define
\[
w(x) = \max(\mathrm{sim}(e(x), q), 0).
\]
We use \(w(x)\) exclusively for scalar relevance weights. For projection-based methods, \(P_C(\cdot)\) denotes orthogonal projection onto the span of \(\{e(c): c\in C\}\). When used in the DPP kernel, we interpret similarity as the inner product of \(\ell_2\)-normalized embeddings, i.e., \(\mathrm{sim}(e(i),e(j))=\langle \tilde e(i), \tilde e(j)\rangle\), so the resulting kernel is a weighted Gram matrix.




\subsection{Compared Strategies}

We compare pruning strategies ranging from lightweight heuristics to LLM-based and learned methods, each instantiating a different notion of marginal value.

\subsubsection*{Heuristic Strategies}
Heuristic strategies estimate marginal value through fixed scoring functions requiring no training or inference. We define the following heuristic strategies below.

\paragraph{Maximal Marginal Relevance (MMR)(\S~\ref{sec:mmr}):}

MMR balances query relevance against redundancy with respect to already retained context~\cite{carbonell1998use}:
\begin{multline*}
V_{\mathrm{MMR}}(x\mid C,Q)
= \lambda\,\mathrm{sim}(e(x), q) \\
- (1-\lambda)\max_{c\in C}\mathrm{sim}(e(x), e(c)),
\end{multline*}
where $\lambda \in [0,1]$ controls the relevance--novelty trade-off.

\paragraph{Geometric Residual Novelty (GRN)(\S~\ref{sec:geometric_residual_novelty}):}

GRN measures whether a candidate introduces a new semantic direction relative to the retained context:
\[
V_{\mathrm{GRN}}(x\mid C) = \|e(x) - P_C(e(x))\|_2,
\]
where \(P_C(e(x))\) is the projection of \(e(x)\) onto the subspace spanned by the retained context embeddings.

\paragraph{Centroid Drift (CD)(\S~\ref{sec:centroid_drift}):}
CD measures whether adding a candidate shifts the semantic center of the retained context~\cite{radev2004centroid}. Let $\mu_C = \frac{1}{|C|}\sum_{c \in C} e(c) $. We define
\[
\mathcal{V}_{\mathrm{CD}}(x \mid C)
=
1 - \mathrm{sim}(\mu_C, \mu_{C \cup \{x\}}).
\]
Larger scores indicate larger semantic drift. Since cosine similarity lies in \([-1,1]\), this score lies in \([0,2]\), although empirical values are typically much smaller.

\paragraph{Determinantal Point Processes (DPP)(\S~\ref{sec:dpp}):}

DPPs favor subsets that are both relevant and diverse~\cite{kulesza2012determinantal}. With kernel entries
\[
L_{ij} = w(i)\,w(j)\,\langle \tilde e(i), \tilde e(j)\rangle,
\]
where \(w(i)\) is the nonnegative query-relevance weight defined above and the gain of adding \(x\) is
\[
V_{\mathrm{DPP}}(x\mid C) = \frac{\det(L_{C\cup\{x\}})}{\det(L_C)}.
\]

Redundant candidates contribute less additional volume and receive smaller gains.

\paragraph{Submodular Coverage (SC)(\S~\ref{sec:submodular_coverage}):}
SC measures how much a retained set covers the semantic space of the candidate pool while accounting for token cost~\cite{lin2011class}. With coverage function
\[
F(C) =
\sum_{j\in P}
\max_{i\in C}
\big[w(i)\,\mathrm{sim}(e(i), e(j))\big]
\]
the token-normalized marginal gain is
\[
\mathcal{V}_{\mathrm{SC}}(x \mid C, Q)
=
\frac{F(C \cup \{x\}) - F(C)}{\mathrm{cost}(x)}.
\]

\paragraph{Combined and lexical variants (\S\ref{sec:lexical_pruning}).}
We also study secondary variants that either combine relevance, novelty, and coverage into a single score or replace dense semantic similarity with cheaper lexical proxies such as TF--IDF cosine similarity and bigram overlap. These variants test whether richer signal integration or lower-cost similarity measures improve the trade-off frontier.

We further evaluate a small set of targeted mixed variants (\emph{CD + SC}, \emph{CD + LLM}, and \emph{SC + LLM}) to test whether different pruning objectives are complementary across stages. These are not intended to exhaust all pairings, but to probe whether novelty-sensitive early pruning combines effectively with coverage-aware or semantic late-stage refinement.

\subsubsection*{LLM-based Strategy (\S~\ref{sec:llm_query_pruning})} 
We evaluate methods in which a language model acts as a pruning judge, predicting whether a candidate should be retained given the current query, context, and candidate set. This method can capture richer semantic judgments than fixed heuristics, but it also introduces additional inference cost.

\subsubsection*{Learned Strategy (\S~\ref{sec:learned-pruning})} 

We also study a learned pre-retrieval controller trained on execution trajectories as an exploratory test of whether a lightweight data-driven proxy can approximate a downstream utility signal.

%% file: sections/06_experiments.tex
\section{Experimental Setup}
\label{sec:exp}



\subsection{Benchmark}

We evaluate pruning strategies on \textbf{DeepResearchGym} \cite{deepresearchgym}, an open-source evaluation sandbox for deep research systems. Its evaluation protocol is built on \textbf{1,000} complex, high-engagement queries from the \textit{Researchy Questions} dataset \cite{rosset2024researchyquestionsdatasetmultiperspective}. Because evaluating every configuration on the full benchmark is computationally expensive, we report results on a fixed \textbf{sample of 100 queries}. 

\subsection{Pipeline and Baseline}
\label{sec:pipeline_baseline}

Our experiments are built on the \textit{GPT-Researcher} framework, which performs multi-step query decomposition, retrieval, aggregation, and report generation. The baseline is the standard \textit{GPT-Researcher} pipeline \emph{without any explicit pruning policy}, including the framework's default late-stage context trimming before synthesis. All pruning variants are implemented on top of the same underlying search policy, and where applicable, we reuse cached subqueries and retrieved evidence across runs, so observed differences are attributable to pruning decisions under this fixed execution framework. This design isolates the effect of pruning stage and scoring rule, but does not fully separate adaptive pruning from a static reduction in search budget; a matched-budget shallow-search baseline is therefore an important control for future work.

\subsection{Compared Configurations}

We compare pruning strategies at three intervention points in the pipeline: \emph{Pre-Retrieval}, \emph{Post-Retrieval}, and \emph{Pre-Synthesis}. We evaluate one-stage policies, multi-stage combinations, heuristic criteria, LLM-based pruning, and learned pruning under a shared execution framework. Unless otherwise specified, all methods use the same underlying generator and report-writing configuration, and differ only in how and where pruning is applied. Unless otherwise noted, the main paper reports one representative operating point per method rather than exhaustively tuning each pruning family to its best possible frontier. Appendix~\ref{sec:threshold_sensitivity} reports local threshold sweeps for five representative post-retrieval methods and shows that the published settings lie within stable sampled operating regions under a 2\% quality-degradation criterion.

\subsection{Evaluation Metrics}


We evaluate methods along four dimensions: \emph{quality}, \emph{relevance},
\emph{faithfulness/source grounding}, and \emph{efficiency}.

\begin{itemize}

    \item \textbf{Quality} (\S~\ref{sec:qual_rel_faith}): Measured with a fixed rubric-based LLM judge applied uniformly across all methods. Because absolute rubric scores can shift across judges, we interpret these values primarily as \emph{relative comparisons under a fixed independent judge}. Appendix Table~\ref{tab:judge_sensitivity} reports a judge-sensitivity analysis.

    \item \textbf{Relevance} (\S~\ref{sec:qual_rel_faith}): KPR+KPC (Key Point Recall + Key Point Contradiction), measuring coverage of key informational points from ground-truth documents while penalizing conflicting statements.

    \item \textbf{Source grounding} (\S~\ref{sec:qual_rel_faith}): Citation Recall (Cit.), an evidence-retention proxy measuring the fraction of report claims linked to a retrievable source. This metric does not by itself establish factual correctness.

    \item \textbf{Efficiency} (\S~\ref{sec:num_nodes}): Node count (Nodes), total token cost in thousands (Tokens), and wall-clock runtime (Time).

\end{itemize}

All metrics are mean values across 100 reports. Formal definitions are in Appendix~\ref{sec:metric_computation}. Because long-horizon research runs are expensive, we evaluate all methods on the same fixed 100-query subset and interpret results as paired comparisons within this setting. We treat small judged-quality differences cautiously: the strongest evidence is that several pruning configurations substantially reduce nodes, tokens, and runtime while approximately preserving judged quality. Appendix~\ref{sec:judge_robustness} reports judge sensitivity, and Appendix~\ref{sec:threshold_sensitivity} reports local threshold sweeps.

\section{Results and Discussion}
\label{sec:results_analysis}

We next compare pruning strategies under the shared evaluation setup above, varying both the pruning stage and the pruning rule while keeping the underlying GPT-Researcher pipeline fixed. This setting isolates the effect of pruning decisions from changes to generation or retrieval. The following subsections examine in detail which pruning stages are most effective, which heuristics work best at each stage, and how quality, efficiency, relevance, and faithfulness trade off across one-, two-, and three-stage settings.

\subsection{One-Stage Pruning: Post-Retrieval vs. Pre-Synthesis}

\begin{table}[t]
\centering
\caption{One-stage pruning results. Full metric breakdown appears in the appendix. Tokens are reported in thousands (k). Bold denotes the best value among pruned methods within each stage block.}
\label{tab:quality_one_stage_main}
\scriptsize
\setlength{\tabcolsep}{3pt}
\renewcommand{\arraystretch}{1.08}
\resizebox{\columnwidth}{!}{
\begin{tabular}{l c c c c c c}
\toprule
\textbf{Method} & \textbf{Nodes} & \textbf{Tokens} & \textbf{Time} & \textbf{Ov.Quality} & \textbf{KPR+KPC} & \textbf{Cit.} \\
\midrule
Baseline & 29.0 & 375.4 & 3422.6 & 57.83 & 70.23 & 95.54 \\
\midrule
\multicolumn{7}{l}{\textbf{Post-Retrieval Pruning}} \\
MMR (\S\ref{sec:mmr}) & \textbf{8.84} & \textbf{114.6} & \textbf{1379.8} & 56.62 & 63.49 & 91.70 \\
GRN (\S\ref{sec:geometric_residual_novelty}) & 13.31 & 175.7 & 2087.9 & 57.02 & 42.89 & 94.40 \\
CD (\S\ref{sec:centroid_drift}) & 10.47 & 137.5 & 1605.7 & 56.84 & 41.29 & 92.48 \\
SC (\S\ref{sec:submodular_coverage}) & 10.82 & 141.9 & 1737.2 & 55.22 & 47.94 & 91.97 \\
DPP (\S\ref{sec:dpp}) & 9.88 & 129.6 & 1520.6 & 54.51 & 43.33 & \textbf{95.62} \\
Combined (\S\ref{sec:lexical_pruning}) & 9.00 & 117.8 & 1399.5 & 56.03 & 64.13 & 92.59 \\
Hybrid (\S\ref{sec:lexical_pruning}) & 9.28 & 121.3 & 1449.5 & 57.63 & \textbf{65.23} & 92.16 \\
LLM (\S\ref{sec:llm_query_pruning}) & 14.90 & 211.8 & 2310.7 & \textbf{59.65} & 49.65 & 93.54 \\

\midrule
\multicolumn{7}{l}{\textbf{Pre-Synthesis Pruning}} \\
MMR (\S\ref{sec:mmr}) & 29.0 & 366.0 & 4446.9 & 57.77 & 65.26 & 90.41 \\
GRN (\S\ref{sec:geometric_residual_novelty}) & 29.0 & 374.1 & 4493.3 & 55.98 & 41.67 & 90.38 \\
CD (\S\ref{sec:centroid_drift}) & 29.0 & 374.5 & 4562.4 & 52.38 & 43.75 & 90.19 \\
SC (\S\ref{sec:submodular_coverage}) & 29.0 & 384.7 & 4533.3 & 53.22 & 42.09 & 90.78 \\
DPP (\S\ref{sec:dpp}) & 29.0 & 374.4 & 4512.4 & 54.70 & 54.07 & 90.24 \\
Combined (\S\ref{sec:lexical_pruning}) & 29.0 & 374.5 & 4408.6 & 59.38 & \textbf{66.32} & 93.79 \\
Hybrid (\S\ref{sec:lexical_pruning}) & 29.0 & \textbf{332.3} & \textbf{3834.1} & \textbf{60.68} & 65.62 & \textbf{95.07} \\
LLM (\S\ref{sec:llm_query_pruning}) & 29.0 & 386.7 & 4512.0 & 57.17 & 44.47 & 92.43 \\
\bottomrule

\end{tabular}}
\end{table}

\paragraph{Best quality vs.\ best efficiency.}

Table~\ref{tab:quality_one_stage_main} shows a sharp contrast between the strongest quality point and the strongest efficiency point. Under the fixed rubric-based LLM judge used in the main experiments, the highest one-stage quality is \emph{Pre-Synthesis Hybrid} at 60.68, improving on the baseline (57.83) by +2.85 points. Its efficiency gains are modest, however: token usage falls only to 332.3k and runtime remains high at 3834.1s, confirming that late pruning can refine the final synthesis context but cannot recover most upstream search cost. By contrast, the strongest efficiency point is \emph{Post-Retrieval MMR}, which reduces token cost to 114.6k, explored nodes to 8.84, and runtime to 1379.8s. Relative to the baseline, this is roughly 69.5\% lower token usage and 59.7\% lower runtime while retaining 56.62 overall quality, or about 97.9\% of baseline quality. This difference is intuitive: \emph{MMR} is most effective early, where its relevance--redundancy trade-off prevents low-value branch expansion before retrieval and result-processing costs are incurred, whereas \emph{Hybrid} is more useful as a late-stage quality refinement mechanism.

\paragraph{Faithfulness and evidence retention.}
The best one-stage citation recall among pruned methods is \emph{Post-Retrieval DPP} at 95.62, whereas \emph{Pre-Synthesis Hybrid} achieves the highest overall quality. This mismatch illustrates a central trade-off in the paper: stronger final report quality does not necessarily imply stronger evidence retention. More importantly, \textbf{no pruned one-stage method surpasses the baseline on KPR+KPC}, indicating that compression can preserve or improve report quality while still discarding evidence needed for maximal relevance retention. Among the post-retrieval heuristics, \emph{DPP} remains more competitive on citation recall because its diversity objective favors retaining a broader evidence set, whereas \emph{MMR} is more aggressive in removing semantically overlapping content and therefore delivers stronger efficiency gains.

\subsection{Two-Stage Pruning: Post-Retrieval + Pre-Synthesis}

\begin{table}[t]
\centering
\caption{Two-stage pruning results. The full metric breakdown appears in Appendix Table~\ref{tab:quality_all}. Tokens are reported in thousands (k). Bold denotes the best value within each subsection among pruned methods. A single method name indicates that the same pruning rule is used at both \emph{Post-Retrieval} and \emph{Pre-Synthesis}; ``A + B'' indicates method A at \emph{Post-Retrieval} and method B at \emph{Pre-Synthesis}.}

\label{tab:quality_two_stage_main}
\scriptsize
\setlength{\tabcolsep}{3pt}
\renewcommand{\arraystretch}{1.08}
\resizebox{\columnwidth}{!}{
\begin{tabular}{l c c c c c c}
\toprule
\textbf{Method} & \textbf{Nodes} & \textbf{Tokens} & \textbf{Time} & \textbf{Ov.Quality} & \textbf{KPR+KPC} & \textbf{Cit.} \\
\midrule
Baseline & 29.0 & 375.4 & 3422.6 & 57.83 & 70.23 & 95.54 \\
\midrule
\multicolumn{7}{l}{\textbf{Main heuristic families}} \\
MMR (\S\ref{sec:mmr}) & \textbf{8.84} & \textbf{114.6} & \textbf{1381.3} & 56.40 & \textbf{65.16} & 92.61 \\
GRN (\S\ref{sec:geometric_residual_novelty}) & 13.2 & 175.2 & 2080.2 & 57.94 & 64.47 & 93.06 \\
CD (\S\ref{sec:centroid_drift}) & 9.02 & 138.2 & 1619.8 & 57.44 & 43.23 & 92.94 \\
SC (\S\ref{sec:submodular_coverage}) & 10.80 & 142.1 & 1776.5 & 57.00 & 62.23 & \textbf{94.74} \\
DPP (\S\ref{sec:dpp}) & 9.2 & 122.8 & 1462.3 & 56.73 & 62.33 & 93.62 \\
LLM (\S\ref{sec:llm_query_pruning}) & 15.57 & 226.6 & 2632.3 & \textbf{58.27} & 48.61 & 93.01 \\
\midrule
\multicolumn{7}{l}{\textbf{Mixed variants}} \\
Combined (\S\ref{sec:lexical_pruning}) & \textbf{8.98} & \textbf{117.6} & \textbf{1392.2} & 54.87 & \textbf{64.88} & 93.03 \\
Hybrid (\S\ref{sec:lexical_pruning}) & 9.30 & 121.7 & 1457.4 & 56.92 & 64.25 & 93.51 \\
CD + SC (\S\ref{sec:lexical_pruning}) & 10.45 & 137.5 & 1599.6 & \textbf{59.47} & 44.29 & 89.96 \\
CD + LLM (\S\ref{sec:lexical_pruning}) & 10.24 & 136.0 & 1589.3 & 58.65 & 63.40 & \textbf{94.34} \\
SC + LLM (\S\ref{sec:lexical_pruning}) & 10.54 & 139.1 & 1700.6 & 58.08 & 64.54 & 93.43 \\
\bottomrule

\end{tabular}}
\end{table}

\paragraph{Best quality vs.\ best efficiency.}
Table~\ref{tab:quality_two_stage_main} shows that combining early and late pruning yields the strongest practical trade-offs in the study. Under the fixed rubric-based LLM judge used in the main experiments, the highest quality point among the evaluated two-stage variants is \emph{CD + SC}, which reaches 59.47 overall quality, improving on the baseline by +1.64 points while reducing token cost by 63.4\%, runtime by 53.3\%, and node count from 29.0 to 10.45. This shows that multi-stage pruning can improve report quality while substantially shrinking the search process. The strongest pure efficiency point remains \emph{MMR}, which reduces token cost further to 114.6k, nodes to 8.84, and runtime to 1381.3s, while reaching 56.40 overall quality. The contrast reflects the underlying objectives: early \emph{MMR} is highly effective at removing redundant branches before they expand, whereas \emph{CD + SC} benefits from a later coverage-aware refinement stage that improves final report quality.
\paragraph{Faithfulness and evidence retention.}

The two-stage setting makes the trade-off frontier most visible. Among the main heuristic families, \emph{SC} achieves the strongest citation recall (94.74), whereas \emph{MMR} is strongest on cost and KPR+KPC (65.16) among pruned methods. This suggests that relevance--redundancy control is effective for reducing search cost, while coverage-oriented objectives are better suited to retaining a broader evidence set for source grounding. LLM-augmented variants remain quality-competitive, but their extra inference cost reduces their end-to-end efficiency advantage. The strongest quality point, \emph{CD + SC}, also shows that quality and faithfulness can diverge: it improves overall quality, but its citation recall (89.96) is weaker than both the baseline and \emph{SC}.

\subsection{Three-Stage Pruning: Pre-Retrieval + Post-Retrieval + Pre-Synthesis}

In the three-stage setting, we compare the main heuristic families against a small set of richer or mixed variants that add lexical, learned, or LLM-based pruning components.

\begin{table}[t]
\centering
\caption{Three-stage pruning results. The full metric breakdown appears in Appendix Table~\ref{tab:quality_all}. Tokens are reported in thousands (k). Bold denotes the best value within each subsection among pruned methods. A single method name means the same pruning rule is used at \emph{Pre-Retrieval}, \emph{Post-Retrieval}, and \emph{Pre-Synthesis}; a notation of the form ``A + B + C'' indicates method A at \emph{Pre-Retrieval}, method B at \emph{Post-Retrieval}, and method C at \emph{Pre-Synthesis}.}
\label{tab:quality_three_stage_main}
\scriptsize
\setlength{\tabcolsep}{3pt}
\renewcommand{\arraystretch}{1.08}
\resizebox{\columnwidth}{!}{
\begin{tabular}{l c c c c c c}
\toprule
\textbf{Method} & \textbf{Nodes} & \textbf{Tokens} & \textbf{Time} & \textbf{Ov.Quality} & \textbf{KPR+KPC} & \textbf{Cit.} \\
\midrule
Baseline & 29.0 & 375.4 & 3422.6 & 57.83 & 70.23 & 95.54 \\
\midrule
\multicolumn{7}{l}{\textbf{Main heuristic families}} \\
MMR (\S\ref{sec:mmr}) & \textbf{7.82} & \textbf{100.1} & \textbf{1157.7} & 55.90 & 63.43 & 91.84 \\
GRN (\S\ref{sec:geometric_residual_novelty}) & 11.57 & 150.0 & 1937.7 & 56.68 & 45.13 & 91.33 \\
CD (\S\ref{sec:centroid_drift}) & 9.27 & 120.6 & 1669.8 & 56.00 & 43.64 & 93.30 \\
SC (\S\ref{sec:submodular_coverage}) & 9.35 & 121.8 & 1438.9 & 55.52 & \textbf{65.79} & 92.03 \\
DPP (\S\ref{sec:dpp}) & 8.68 & 113.3 & 1626.2 & 55.41 & 46.33 & \textbf{95.40} \\
LLM (\S\ref{sec:llm_query_pruning}) & 10.12 & 143.6 & 1574.7 & \textbf{59.53} & 65.09 & 93.12 \\
\midrule
\multicolumn{7}{l}{\textbf{Mixed variants}} \\
Lexical + CD + SC (\S\ref{sec:lexical_pruning}) & 9.26 & 120.1 & 1375.0 & 57.81 & 42.67 & 93.31 \\
Combined (\S\ref{sec:lexical_pruning}) & \textbf{7.90} & \textbf{102.8} & \textbf{1227.6} & 55.00 & 62.50 & 90.83 \\
Hybrid (\S\ref{sec:lexical_pruning}) & 8.16 & 106.3 & 1282.7 & 57.10 & \textbf{64.80} & 93.06 \\
Learned Query + GRN + GRN (\S\ref{sec:learned-pruning}) & 10.44 & 145.7 & 1744.4 & \textbf{58.13} & 51.78 & \textbf{95.48} \\
\bottomrule

\end{tabular}}
\end{table}

\paragraph{Best quality vs.\ best efficiency.}

Table~\ref{tab:quality_three_stage_main} shows that three-stage pruning is most useful when the objective is maximal compression. Among the main heuristic families, \emph{MMR} is the strongest compression point, reducing token cost to 100.1k, explored nodes to 7.82, and runtime to 1157.7s, or roughly 73.3\% lower token usage than the baseline. Its overall quality (55.90) remains below both the baseline and the strongest quality-oriented variants, indicating that extending relevance--redundancy pruning across all three stages sharpens compression more than it improves end-task quality.

\paragraph{Faithfulness and evidence retention.}

Under the fixed rubric-based LLM judge used in the main experiments, the highest overall quality in this family is achieved by \emph{LLM} at 59.53, while \emph{Learned Query + GRN} attains the strongest citation recall (95.48) and \emph{SC} the strongest KPR+KPC (65.79) among pruned methods. Adding a third stage therefore does not remove the trade-off between quality, compression, and evidence retention; it makes it more explicit. Relative to the strongest two-stage configurations, three-stage pruning improves compression more reliably than quality, making the third stage most valuable when aggressive cost reduction is prioritized over maximal end-task quality.

\subsection{Cross-Stage Findings}

\paragraph{How do pruning objectives differ?}
Across settings, methods based on explicit relevance--redundancy control, especially \emph{MMR}, are the strongest compression tools. Their advantage is greatest when applied early, where pruning a redundant branch avoids not only final-context growth but also the downstream cost of retrieval, result processing, and recursive expansion. By contrast, coverage-oriented methods such as \emph{SC} are less aggressive compressors but often stronger on evidence retention because they reward keeping a broader set of complementary support rather than eliminating overlap as aggressively as possible. Diversity-oriented methods such as \emph{DPP} behave similarly, favoring a wider evidence set and therefore remaining more competitive on citation recall than on pure efficiency. Geometric novelty methods such as \emph{GRN} preserve novel semantic directions but are less tightly coupled to aggressive cost reduction. Centroid-based methods such as \emph{CD} lie between these extremes: they capture novelty through shifts in the semantic center of the retained context, helping broader contextual refinement, but with a coarser signal than direct relevance--redundancy control and therefore a weaker alignment with aggressive efficiency gains.

\paragraph{How do methods change across stages?}
Holding the pruning rule fixed reveals a second consistent pattern. For \emph{MMR}, moving from one-stage to two-stage leaves the strongest efficiency point essentially unchanged (\textbf{114.6k} tokens and \textbf{8.84} nodes in both settings), while three-stage pruning pushes compression further, reducing token cost to \textbf{100.1k} and nodes to \textbf{7.82}, but with weaker overall quality. This suggests that \emph{MMR} captures most of its benefit once applied at \emph{Post-Retrieval}, and that additional stages mainly sharpen compression rather than improve end-task quality. By contrast, \emph{SC} benefits less from extra stages on pure efficiency, but remains comparatively stronger on evidence-retention objectives such as citation recall and KPR+KPC. More broadly, additional stages improve cost reduction more reliably than report quality, whereas later-stage refinement is most useful when the objective places greater weight on final synthesis quality than on maximal efficiency.

\paragraph{Cross-benchmark evidence.}
Although our primary analysis is conducted on DeepResearchGym, supplementary results on DeepResearch Bench (Appendix~\ref{sec:deepbench}, Tables~\ref{tab:bench_quality_summary} and~\ref{tab:bench_efficiency_summary}) show the same broad efficiency pattern: earlier pruning yields substantially larger token and runtime savings than root-only pruning, and \emph{MMR} remains one of the strongest compression-oriented heuristics. At the same time, method-level quality rankings are less stable across the two benchmarks (Table~\ref{tab:bench_quality_summary}). We therefore interpret these results as directional support for the stage-ordering conclusion rather than as a full replication of the quality trade-offs observed on DeepResearchGym, and note that the two benchmarks use different evaluation protocols and report-quality metrics.

\paragraph{Which stages and configurations matter most?}
Comparing stages and configurations head-to-head, \emph{Post-Retrieval} pruning delivers the largest efficiency gains while remaining relatively close to baseline quality, because it prevents low-value branches from expanding before retrieval and result-processing costs are incurred. By contrast, one-stage \emph{Pre-Synthesis} pruning gives the strongest quality refinement, but is usually too late to recover most upstream cost. At the configuration level, two-stage pruning offers the strongest practical quality-efficiency trade-off by combining early search control with late refinement. Three-stage pruning pushes compression further, but with smaller returns for report quality, making it most useful when the objective is maximal cost reduction rather than maximal end-task quality.

\paragraph{When do Mixed variant methods help?}
Mixed variant and later-stage pruning rules become more competitive when the target is final report quality rather than absolute efficiency. \emph{Pre-Synthesis Hybrid} gives the strongest one-stage quality, and LLM-based variants are often quality-competitive because they can make more flexible semantic decisions than fixed heuristics. However, their additional inference cost limits their efficiency advantage, making them most attractive when quality is prioritized over absolute cost.

\paragraph{What the exploratory learned controller suggests.}
The learned pre-retrieval controller serves as a small-scale data-driven alternative to hand-designed marginal-value criteria. In our results, it produces viable operating points in a limited number of settings, especially when predictive query filtering is useful, but it does not provide strong evidence of consistent superiority over the strongest heuristic baselines. This is informative in itself: most of the attainable gains in our study come from a well-matched pruning objective and stage placement, while the learned component is best viewed as an exploratory proof of concept rather than a mature replacement for simpler heuristics.

\section{Conclusion}
\label{sec:conclusion}

We presented the first systematic stage-aware study of marginal-value-based pruning for long-horizon deep research agents. Across \emph{Pre-Retrieval}, \emph{Post-Retrieval}, and \emph{Pre-Synthesis}, stage placement is often more consequential than the scoring rule: early pruning, especially at \emph{Post-Retrieval}, yields the largest savings in nodes, tokens, and runtime, whereas later pruning mainly refines the final synthesis context. No single strategy dominates across objectives. Under the fixed rubric-based judge used in the main experiments, \emph{Pre-Synthesis Hybrid} gives the strongest one-stage quality, \emph{CD + SC} the strongest observed quality-efficiency trade-off, and three-stage pruning is most useful for maximal compression. Learned pruning remains exploratory; most gains come from careful stage placement and lightweight heuristics. Because report quality and evidence retention can diverge under pruning, policies should be evaluated jointly in terms of quality, efficiency, relevance, and faithfulness. Overall, our results support a stage-aware view of pruning and offer practical guidance for efficient long-horizon research agents.

\section{Limitations}
\label{sec:limitations}

This study has several limitations. First, our empirical analysis is conducted within a fixed deep research pipeline and benchmark setup, so the exact trade-offs may change under different agent architectures, retrieval systems, or underlying language models. Our conclusions are therefore strongest as a comparative study of pruning strategies within this setting rather than as universal claims about all long-horizon agents. Although supplementary experiments on DeepResearch Bench (Appendix~\ref{sec:deepbench}, Tables~\ref{tab:bench_quality_summary} and~\ref{tab:bench_efficiency_summary}) support the broad efficiency advantage of earlier pruning, method-level quality rankings are less stable across benchmarks, suggesting that quality-sensitive conclusions depend more strongly on the evaluation protocol and task formulation than the basic stage-ordering effect. Our quality metric is also judge-dependent. Appendix Table~\ref{tab:judge_sensitivity} shows that absolute rubric scores can shift substantially across judge choices, even for the same generated reports. We therefore interpret quality-sensitive conclusions more cautiously than efficiency results, and treat the reported quality values primarily as relative comparisons under a fixed independent judge rather than as stable absolute measurements. Relatedly, as discussed in Section~\ref{sec:pipeline_baseline}, our comparison does not include a matched-budget shallow-search baseline; thus, the efficiency gains should be interpreted as gains from stage-aware pruning under a fixed search policy, rather than as evidence that pruning dominates all static budget-reduction strategies.

Second, several pruning methods require threshold or hyperparameter choices, and the best operating point remains task-dependent. Appendix~\ref{sec:threshold_sensitivity} reports local threshold sweeps for five representative post-retrieval methods and shows that the published settings lie within stable sampled regions under a 2\% quality-degradation criterion. These sweeps suggest that our main conclusions are not driven by isolated brittle thresholds, although finer-grained method ordering may still shift under different threshold choices or under exhaustive per-method tuning.

Finally, our evaluation metrics do not capture all aspects of utility or reliability. Automatic measures of report quality, relevance, and citation recall only partially reflect factual correctness, completeness of evidence, and usefulness to end users. Our findings should therefore be interpreted as a systematic empirical study of pruning trade-offs rather than a complete evaluation of deployment readiness.

\section{Ethics Statement}
\label{sec:ethics}

Improving the efficiency of long-horizon research agents can reduce computational cost, latency, and environmental burden. However, more efficient systems can also make it easier to scale the production of misleading or low-quality synthesized content.

Pruning introduces an additional risk: imperfect pruning signals may discard contradictory evidence, minority viewpoints, or important caveats while preserving fluent but insufficiently supported summaries. Pruning should therefore not be treated as a substitute for verification. We recommend pairing pruning-based research agents with source transparency, citation auditing, and human oversight, especially in high-stakes domains such as medicine, law, public policy, and education.

Learned pruning models may also inherit biases from the trajectories and evaluation signals used to train them. Their outputs should therefore be understood as objective-driven decisions rather than as ground truth about what evidence is important.

%% file: sections/05_metric_computation.tex
\section{Appendix}
\label{sec:appendix}

This appendix provides supporting detail for the main paper along three dimensions. First, it defines the reported metrics and derived efficiency quantities, including node counts, token accounting, runtime decomposition, and stage-wise pruning effectiveness. Second, it expands the method descriptions, giving the mathematical form and stage-specific role of each pruning strategy summarized in the main text. Third, it reports the full comparison tables and robustness analyses that underlie the compact main-paper results.

The purpose of these materials is to make the empirical findings more transparent rather than to introduce new claims. In particular, the appendix tables expose the full metric breakdown behind the stage-aware comparisons in the main paper, clarify where efficiency gains arise within the pipeline, and show how different pruning strategies behave under more detailed accounting than can fit in the main text.

\subsection{Metric Computation Details}
\label{sec:metric_computation}

All reported cost, runtime, and pruning metrics are aggregated over the same set of evaluation reports. For any per-report quantity $x_i$ measured on report $i \in \{1,\dots,N\}$, we report the sample mean
\[
\bar{x} = \frac{1}{N}\sum_{i=1}^{N} x_i
\]
and, when available, the standard error
\[
\mathrm{SE}(x) = \frac{s_x}{\sqrt{N}}, \qquad
s_x^2 = \frac{1}{N-1}\sum_{i=1}^{N}(x_i - \bar{x})^2.
\]
Unless otherwise stated, tables report values as mean $\pm$ standard error.

\subsection{Quality, relevance, and faithfulness:}
\label{sec:qual_rel_faith}
The quality dimensions reported in the main comparison tables, namely Overall, Clarity, Depth, Balance, Breadth, Support, and Insight, are obtained by running the generated reports through the DeepResearchGym evaluation pipeline and averaging the resulting scores over reports. The same procedure is used for the relevance metric (KPR+KPC) and the faithfulness metric (Citation Recall). These metrics are currently reported as means only.

\subsection{\# Nodes:}
\label{sec:num_nodes}
For the baseline configuration with breadth $4$ and depth $3$, the nominal full search tree contains
\[
1 + 4 + 8 + 16 = 29
\]
nodes, consisting of one planning node and $28$ research nodes. We report the node count computed as
\[
\mathrm{N}_i^{\mathrm{corr}} = 1 + R_i,
\]
where $R_i$ is the number of result-processing calls in report $i$, and the leading $1$ accounts for the planning node. We also report the average number of pruned nodes, denoted $\bar{P}$, from the node-level pruning logs. The pruning rate shown in the appendix is then computed from the reported means as
\[
\mathrm{Pruning\ Rate} = \frac{\bar{P}}{\bar{N}^{\mathrm{corr}}}\times 100.
\]

\subsection{Token cost:}
\label{sec:token_cost}
For each report, total token cost is defined as the sum of input and output tokens across all token-tracked pipeline stages:
\[
T_i = T_i^{\mathrm{in}} + T_i^{\mathrm{out}}
      = \sum_{\phi \in \Phi}\left(T_{i,\phi}^{\mathrm{in}} + T_{i,\phi}^{\mathrm{out}}\right),
\]
where $\Phi$ includes the logged token phases such as planning, query generation, pre-retrieval pruning (when the pruning method itself invokes an LLM), result processing, embedding, and other tracked model calls. The main efficiency table reports $\bar{T}$ as \# Tokens. The token-accounting table further decomposes this into average input tokens, average output tokens, and total tokens.

\subsection{Estimated tokens saved by pruning:}

For each pruning stage $s$ in report $i$, let $C_{i,s}^{\mathrm{before}}$ and $C_{i,s}^{\mathrm{after}}$ denote the tokenized context size before and after pruning. The estimated tokens saved at that stage are
\[
\Delta T_{i,s} = C_{i,s}^{\mathrm{before}} - C_{i,s}^{\mathrm{after}}.
\]
The total estimated tokens saved by pruning in report $i$ are then
\[
T_{i}^{\mathrm{saved}} = \sum_{s} \Delta T_{i,s}.
\]

\subsection{Estimated token reduction:}
\label{sec:est_token_reduction}
The estimated pre-pruning token budget of a method is defined as the final observed token cost plus the estimated tokens removed by pruning:
\[
T_i^{\mathrm{before}} = T_i + T_i^{\mathrm{saved}}.
\]
The estimated total token reduction percentage is then
\[
\mathrm{Token\ Reduction}_i =
\frac{T_i^{\mathrm{saved}}}{ T_i^{\mathrm{before}}}\times 100.
\]
The efficiency table reports the mean of this quantity over reports. This metric measures within-method pruning effectiveness, namely how much of the method's own pre-pruning token budget is removed by pruning.

\subsection{Savings vs. Baseline:}
\label{sec:est_save_baseline}
To quantify absolute savings relative to the unpruned baseline, we compare the mean total token count of each method against the baseline mean total token count. The percentage savings relative to baseline are computed as
\[
\mathrm{Savings\ vs.\ Baseline} =
\left(1 - \frac{\bar{T}_{\mathrm{method}}}{\bar{T}_{\mathrm{baseline}}}\right)\times 100.
\]
When reported in absolute units, mean savings vs.\ baseline are expressed in thousands of tokens:
\[
\mathrm{Mean\ Savings\ vs.\ Baseline} =
\frac{\bar{T}_{\mathrm{baseline}} - \bar{T}_{\mathrm{method}}}{1000}.
\]
Thus, this quantity answers how many fewer tokens, in $k$ tokens per report, a method uses on average relative to baseline.

\subsection{Token breakdown and token share by stage.}
\label{sec:token_break_stage}
For each token phase $\phi$, we aggregate stage-wise token usage as
\[
T_{i,\phi} = T_{i,\phi}^{\mathrm{in}} + T_{i,\phi}^{\mathrm{out}}.
\]
The token breakdown table reports $\bar{T}_{\phi}$ for each stage. The token-share table reports the fraction of the total token budget consumed by each stage:
\[
\mathrm{Share}_{i,\phi} = \frac{T_{i,\phi}}{T_i}\times 100.
\]
This decomposition reveals whether a method's cost is dominated by planning, query generation, pruning, result processing, or embedding. In most methods, result processing dominates the total token budget; in methods with explicit LLM-based query pruning, the pruning stage itself also contributes nontrivially.

\subsection{Runtime metrics:}
\label{sec:runtime_metrics}

For each report, total runtime is measured as wall-clock time from the start to the end of the pipeline. The runtime breakdown table reports the mean and standard error of the total runtime and of each logged stage-level latency. Let $L_{i,\psi}$ denote the duration of stage $\psi$ in report $i$, with reported stage values $\bar{L}_{\psi}$.

We define the measured pipeline stages as follows:

\begin{itemize}
    \item \textbf{Total Runtime (s):} The total elapsed wall-clock time for the entire report generation pipeline.
    \item \textbf{Planning:} Measures the time to generate the initial follow-up questions that decompose the user query into high-level research directions. This stage occurs once at the root and seeds all subsequent stages.
    \item \textbf{Query Generation (Query Gen):} Measures the time to convert follow-up questions—both from Planning and any additional questions generated during Result Processing—into candidate queries suitable for retrieval.
    \item \textbf{Query Pruning:} Measures any extra latency incurred when candidate queries are explicitly filtered before retrieval.
    \item \textbf{Research / Scraping:} Measures the retrieval-heavy stage in which the system issues searches, visits sources, and gathers raw evidence for the candidate queries.
    \item \textbf{Result Processing (Result Proc.):} Measures the time spent processing retrieved context into structured learnings, follow-up questions if needed (i.e. max depth not yet reached), and citations. These follow-up questions later feed into Query Generation.
    \item \textbf{Branch Pruning:} Measures the overhead of post-retrieval pruning during recursive branch aggregation, where candidate branches are evaluated and potentially discarded based on incremental utility.
    \item \textbf{Root Pruning:} Measures the cost of pre-synthesis pruning of the final aggregated context before the report is produced.
\end{itemize}

\subsection{Pruning-stage effectiveness:}
\label{sec:pruning_stage_effectiveness}
To characterize where pruning happens within the pipeline, we report stage-wise pruning ratios and stage-wise token reductions at pre-retrieval, post-retrieval, and pre-synthesis levels. For a given stage $s$, let $n_{i,s}^{\mathrm{before}}$ and $n_{i,s}^{\mathrm{after}}$ denote the number of candidate items before and after pruning. The stage-level pruning ratio is
\[
\mathrm{Ratio}_{i,s} =
\frac{n_{i,s}^{\mathrm{before}} - n_{i,s}^{\mathrm{after}}}{n_{i,s}^{\mathrm{before}}}\times 100.
\]
Likewise, if $C_{i,s}^{\mathrm{before}}$ and $C_{i,s}^{\mathrm{after}}$ denote the tokenized context sizes before and after pruning, the corresponding stage-level token reduction is
\[
\mathrm{TokenRed}_{i,s} =
\frac{C_{i,s}^{\mathrm{before}} - C_{i,s}^{\mathrm{after}}}{C_{i,s}^{\mathrm{before}}}\times 100.
\]
For pre-retrieval pruning, the item count refers to candidate queries; for post-retrieval and pre-synthesis pruning, it refers to retained context items. Reporting both quantities is useful because pruning by item count and pruning by token mass are not identical: removing half of the items does not necessarily remove half of the tokens.

\section{Additional Method Details}
\label{sec:appendix_methods}

This section provides additional detail on the pruning strategies compared in the main paper. We first introduce shared notation, then describe each strategy in turn, including its intuition, mathematical form, and stage-specific role in our pipeline.

We adopt the notation from Section~\ref{sec:problem}. Let $e(\cdot) \in \mathbb{R}^d$ denote the embedding function, and let
\[
q = e(Q)
\]
be the embedding of the root query $Q$. A candidate item is denoted by $x$. Depending on the pruning stage, $x$ may represent a generated subquery, a retrieved context item, or a context block considered for final synthesis. We write $C$ for the current retained context, with elements $c_i \in \mathbb{R}^d$.

Our default semantic similarity is cosine similarity:
\[
\mathrm{sim}(u,v) = \frac{u^\top v}{\|u\|\,\|v\|}.
\]
When budget-awareness is needed, we write $\mathrm{cost}(x)$ for the token cost of candidate $x$.

When a formula operates in embedding space, we write \(e(x)\) for the embedding of candidate \(x\) and \(e(c)\) for the embedding of a retained item \(c\in C\). We use
\[
w(x) = \max(\mathrm{sim}(e(x), q), 0)
\]
for the nonnegative scalar query-relevance weight, reserving \(q\) exclusively for the query embedding. For projection-based methods, \(P_C(\cdot)\) denotes orthogonal projection onto the span of \(\{e(c): c\in C\}\).

\subsection{Heuristic Strategies}
\label{sec:heur-strat}

\paragraph{Maximal Marginal Relevance (MMR)}
\label{sec:mmr}

Maximal Marginal Relevance (MMR) is a classical information retrieval criterion that balances query relevance against redundancy with respect to already selected content~\cite{carbonell1998use}. In our setting, the MMR score of candidate $x$ is
\begin{multline*}
V_{\mathrm{MMR}}(x\mid C, Q)
= \lambda\,\mathrm{sim}(e(x), q) \\
- (1-\lambda)\max_{c\in C}\mathrm{sim}(e(x), e(c)).
\end{multline*}
where $\lambda \in [0,1]$ controls the relevance--novelty trade-off.

The first term rewards semantic alignment with the research query, while the second penalizes overlap with previously retained evidence. A candidate receives a high score only if it is both relevant and non-redundant. Because cosine similarity lies in $[-1,1]$, the effective score range is bounded by the mixture weights, shifting toward query relevance as $\lambda$ increases.

Operationally, MMR is used for \emph{Post-Retrieval} pruning and \emph{Pre-Synthesis} pruning. At \emph{Post-Retrieval}, the score is used as an information-gain test for deciding whether a branch should be retained. At \emph{Pre-Synthesis}, MMR is used greedily to construct a compact final context under a budget by repeatedly selecting the item with the highest marginal MMR score.

\paragraph{Geometric Residual Novelty}
\label{sec:geometric_residual_novelty}

Geometric Residual Novelty (GRN) measures whether a candidate introduces a new semantic direction relative to the retained context. The idea is inspired by residual-based subspace selection methods such as orthogonal matching pursuit~\cite{tropp2007signal}. Let $\mathcal{U}(C)$ denote the subspace spanned by the retained context embeddings, and let $P_C(e(x))$ denote the orthogonal projection of e(x) onto that subspace.. We define the residual
\[
\operatorname{res}(x;C) = e(x) - P_C(e(x)),
\]
and the gain
\[
V_{\mathrm{GRN}}(x\mid C) = \|\operatorname{res}(x;C)\|_2
\]

If the residual norm is small, then $x$ lies largely within the span of existing evidence and contributes little novelty. If the residual norm is large, then $x$ contributes a new semantic direction. With unit-normalized embeddings, $\mathcal{V}_{\mathrm{GRN}} \in [0,1]$.

In our pipeline, GRN is used at \emph{Post-Retrieval} to determine whether a newly retrieved branch adds genuinely new information relative to already accepted branches, and at \emph{Pre-Synthesis} to retain context items that maximize orthogonal novelty under a token or word budget.

\paragraph{Centroid Drift}
\label{sec:centroid_drift}

Centroid drift is an embedding-space adaptation of centroid-based representativeness ideas from multi-document summarization~\cite{radev2004centroid}. Rather than measuring redundancy through only pairwise similarity, it asks whether adding a candidate changes the semantic center of the retained context.

Let
\[
\mu_C = \frac{1}{|C|}\sum_{c\in C} e(c)
\]
denote the centroid of the current context. After adding candidate $x$, the updated centroid becomes
\[
\mu_{C\cup\{x\}} = \frac{1}{|C|+1}\left(\sum_{c\in C} e(c) + e(x)\right).
\]
We define gain as
\[
\mathcal{V}_{\mathrm{CD}}(x \mid C)
=
1 - \mathrm{sim}(\mu_C, \mu_{C \cup \{x\}}).
\]

If adding $x$ barely changes the centroid, then it is already well represented by the retained context and contributes little marginal value. If it produces a large drift, then it expands the semantic footprint of the context. Since cosine similarity lies in $[-1,1]$, the score lies in $[0,2]$, although in practice values are usually much smaller.

Centroid drift is used for \emph{Post-Retrieval} pruning and \emph{Pre-Synthesis} pruning. At \emph{Post-Retrieval}, a branch is retained only if it shifts the semantic center of the accepted evidence enough to exceed a threshold. At \emph{Pre-Synthesis}, we greedily select context items that most increase the semantic footprint of the final retained set.

\paragraph{Determinantal Point Processes (DPP)}
\label{sec:dpp}

Determinantal Point Processes (DPPs) provide a probabilistic framework for selecting subsets that are both individually relevant and mutually diverse~\cite{kulesza2012determinantal}. Let $L$ be a positive semidefinite kernel matrix over candidate items, with entries

\[L_{ij} = w(i)\,w(j)\,\langle \tilde e(i), \tilde e(j)\rangle,
\]
where \(w(i)\) is the nonnegative query-relevance weight defined above.

Because \(L\) is constructed as a weighted Gram matrix over normalized embeddings, it is positive semidefinite by construction.

The gain of adding candidate $x$ to context $C$ is
\[
\mathcal{V}_{\mathrm{DPP}}(x \mid C)
=
\frac{\det(L_{C \cup \{x\}})}{\det(L_C)}.
\]

Geometrically, this ratio measures how much additional volume the candidate contributes to the selected set. Highly redundant candidates contribute little new volume and therefore receive small gains. Because $L$ is positive semidefinite, the determinant ratio is nonnegative, with larger values indicating jointly relevant and diverse candidates.

We use DPP-style gain for \emph{Post-Retrieval} pruning and \emph{Pre-Synthesis} pruning. At \emph{Post-Retrieval}, the determinant ratio acts as an information-gain score for deciding whether a branch is worth retaining. At \emph{Pre-Synthesis}, it is used greedily to construct a diverse final synthesis context.

\paragraph{Submodular Coverage}
\label{sec:submodular_coverage}

Submodular coverage is our most general budget-aware objective. It is based on query-focused facility-location style selection~\cite{lin2011class}, and aims to retain a set of items that collectively covers the semantic space of the candidate pool while accounting for token cost.

 We define the coverage function
\[
F(C) =
\sum_{j\in P}
\max_{i\in C}
\big[w(i)\,\mathrm{sim}(e(i), e(j))\big].
\]

Here, \(P\) denotes the full candidate pool available at the current pruning decision. It is fixed while scoring candidates for that decision step, and is recomputed only when the pipeline advances to a new pruning state. For example, if the current candidate pool is \(P=\{x_1,x_2,x_3\}\) and the retained set is \(C=\{x_1\}\), then the marginal gain of adding \(x_2\) is computed as \(\Delta F(x_2\mid C)=F(\{x_1,x_2\})-F(\{x_1\})\) with coverage still evaluated over the same fixed pool \(P\). Also, \(i\) and \(j\) index candidate items, while \(e(i)\) and \(e(j)\) denote their embeddings. The marginal gain of adding item $x$ is
\[
\Delta_F(x \mid C)
=
F(C \cup \{x\}) - F(C).
\]
To incorporate budget awareness, we normalize by token cost:
\[
\mathcal{V}_{\mathrm{SC}}(x \mid C, Q)
=
\frac{\Delta_F(x \mid C)}{\mathrm{cost}(x)}.
\]

This objective favors candidates that cover large uncovered regions of the pool while remaining economical in token usage. Because redundant items add little new coverage, they receive low gain. Since the underlying objective is monotone submodular, greedy selection provides a natural and efficient approximation procedure.

Submodular coverage is used in our experiments at all three stages. At \emph{Pre-Retrieval}, it scores generated subqueries before search. At \emph{Post-Retrieval}, it prunes low-value retrieved branches. At \emph{Pre-Synthesis}, it compresses the final context under a budget.

\paragraph{Combined Scoring}
\label{sec:combined_scoring}

Because relevance, novelty, and coverage capture different aspects of marginal value, we also study combined scoring functions that integrate multiple signals. A generic combined score can be written as
\begin{multline*}
\mathrm{Score}(x)
=
\alpha\,\mathrm{Rel}(x,Q) \\
+ \beta\,\mathrm{Nov}(x,C)
+ \gamma\,\mathrm{Cov}(x,C,Q),
\end{multline*}
where $\alpha,\beta,\gamma \ge 0$ are mixture weights.

Our concrete hybrid instantiation combines direct query relevance, geometric novelty, and submodular coverage:
\begin{multline*}
\mathrm{Score}(x) =
\alpha \, \mathrm{sim}(e(x), q) \\
- \beta \, V_{\mathrm{GRN}}(x \mid C)
+ \gamma \, V_{\mathrm{SC}}(x \mid C, Q)
\end{multline*}
with $\alpha+\beta+\gamma = 1$.

In the main Hybrid configuration used in our experiments, we set \((\alpha,\beta,\gamma)=(0.40,0.30,0.30)\).

The first term rewards direct alignment with the research goal, the second rewards orthogonal novelty, and the third rewards global coverage under a token-aware objective. We use this hybrid to study whether explicitly combining local relevance, novelty, and global coverage yields a better quality-efficiency trade-off than any single criterion alone.

\paragraph{Lexical Pruning}
\label{sec:lexical_pruning}

We additionally study a lexical pruning variant that replaces dense embedding similarity with a lightweight surface-form proxy. Each text is represented by a lexical profile consisting of TF--IDF weights and bigram sets. Similarity between two items is defined as a weighted combination of TF--IDF cosine similarity~\cite{ramos2003using} and bigram Jaccard overlap~\cite{broder1997resemblance}:
\begin{multline*}
\mathrm{LexSim}(x,y)
=
w_{\mathrm{tfidf}}\,\mathrm{CosSim}_{\mathrm{tfidf}}(x,y) \\
+
w_{\mathrm{jac}}\,\mathrm{Jaccard}_{\mathrm{bigram}}(x,y),
\end{multline*}
with $w_{\mathrm{tfidf}} + w_{\mathrm{jac}} = 1$.

Using this lexical similarity, we define a lexical MMR-style score
\begin{multline*}
\mathcal{V}_{\mathrm{Lex}}(x \mid C,Q)
=
\lambda\,\mathrm{LexSim}(x,Q) \\
-
(1-\lambda)\max_{y \in C}\mathrm{LexSim}(x,y).
\end{multline*}

This criterion preserves the same relevance-versus-redundancy structure as embedding-based MMR, but avoids the cost of dense embedding computation. We use it as a cheap baseline for studying how much pruning performance depends on richer semantic representations.

\subsection{LLM-Based Pruning}
\label{sec:llm_query_pruning}

We evaluate LLM-based pruning variants in which a language model acts as a pruning judge. Given the current query, context, and candidate set, the model is prompted to assess the usefulness of each candidate and return structured outputs such as predicted gain, keep probability, or a keep-versus-prune decision. The system then retains the highest-scoring candidates or candidates above a threshold, depending on the stage.

Compared with fixed heuristic rules, this approach can incorporate richer semantic judgments and broader task context. However, it also introduces additional inference cost because pruning itself requires one or more model calls. In our experiments, LLM-based pruning is evaluated both as a standalone policy and in mixed-stage combinations with heuristic downstream pruning.

\subsection{Learning-Based Pruning}
\label{sec:learned-pruning}

We also study a learned controller for \emph{Pre-Retrieval} branch selection. Its role is intentionally narrow: given the current research state and a set of candidate subqueries, it predicts which candidate branches are most worth expanding before retrieval and downstream processing costs are incurred.

To train this controller, we reconstruct a \emph{candidate-level} supervision dataset from raw generation logs. Each example corresponds to one candidate branch considered at one decision point, together with the local search state available when that decision is made. Across 34 complete runs, this yields \textbf{360} candidate branch decisions. This count is larger than the number of executed branches because each decision point may contain multiple candidate subqueries, only some of which are ultimately expanded. For consistency of targets, we derive labels only from the submodular-family runs (\texttt{GPTResearcher\_sc}), so the learned controller should be interpreted as a \emph{submodular-aligned pre-retrieval proxy} rather than as a fully general learned pruning policy. We also partition the candidate-level dataset by run, so that decisions from the same execution trajectory do not appear in both training and evaluation splits.

Our model is a lightweight multitask neural value model built on the pretrained encoder \textit{BAAI/bge-small-en-v1.5}. One tower encodes the current search state, including the root query, local frame context, parent research goal, and previously retained branches, while a second tower encodes the candidate branch through its query and research goal. We keep the encoder frozen and feed the resulting semantic representations, together with structured search features such as depth, branch order, and prior retained count, into a small MLP. The model has two heads: a regression head that predicts a continuous utility score $\hat{g}(x)$ and a classification head that predicts a keep probability $p_{\mathrm{keep}}(x)$.

At inference time, candidate $x$ is retained if
\[
p_{\mathrm{keep}}(x) \ge \tau,
\]
where $\tau$ is the pruning threshold. In our experiments, we evaluate multiple thresholds, including $\tau = 0.5$ and $\tau = 0.7$, to control pruning aggressiveness. If no candidate exceeds the threshold, we retain the top-scoring candidate as a fallback to avoid degenerate empty expansions.

In the full pipeline, learned \emph{Pre-Retrieval} pruning is paired with submodular \emph{Post-Retrieval} pruning and centroid-drift \emph{Pre-Synthesis} pruning. Because the supervision set remains modest in size after restricting labels to a single target definition, we interpret the learned controller cautiously and treat it primarily as a proof-of-concept pre-retrieval proxy rather than as strong evidence that learned pruning currently outperforms well-designed heuristics. 

\subsection{Stage-Specific Application}
\label{sec:how_used_in_pipeline_appendix}

Different pruning strategies are applied to different candidate types depending on the stage:

\begin{itemize}
    \item \textbf{Pre-Retrieval:} candidate items are generated subqueries, represented by their query text and associated research goals. The objective is to avoid launching redundant or low-value search branches before retrieval cost is incurred.
    \item \textbf{Post-Retrieval:} candidate items are newly retrieved evidence blocks, evaluated against previously retained context. The objective is to stop low-value recursive expansion early.
    \item \textbf{Pre-Synthesis:} candidate items are the final aggregated context blocks. The objective is to construct a compact, diverse, and query-relevant synthesis context under a budget.
\end{itemize}

Operationally, \emph{Post-Retrieval} pruning is typically used as a thresholded gating rule, while \emph{Pre-Retrieval} and \emph{Pre-Synthesis} often involve greedy selection under item or word budgets. This shared formulation lets us compare different notions of marginal value within a common stage-aware execution framework.

\subsection{Supplementary Cross-Benchmark Results on DeepResearch Bench}
\label{sec:deepbench}

To assess whether our main efficiency findings transfer beyond DeepResearchGym, we also evaluate a subset of pruning configurations on DeepResearch Bench. We treat these experiments as supplementary cross-benchmark evidence rather than as a second primary benchmark. In particular, we use them to test whether the stage-ordering conclusion that earlier pruning yields substantially larger end-to-end savings than late-only pruning remains stable under a different evaluation setup. Table~\ref{tab:bench_quality_summary} summarizes overall report quality, and Table~\ref{tab:bench_efficiency_summary} summarizes token and runtime efficiency.

The DeepResearch Bench results support this efficiency trend: branch-only and two-stage pruning consistently yield much larger token and runtime savings than root-only pruning, and \emph{MMR} remains one of the strongest compression-oriented heuristics. However, the method-level quality rankings are less aligned with those observed on DeepResearchGym. We interpret this discrepancy cautiously, as the two benchmarks differ both in task formulation and in how report quality is evaluated: DeepResearchGym uses an LLM-as-a-judge rubric over open-ended reports, whereas DeepResearch Bench reports RACE-style quality metrics. We therefore treat the DeepResearch Bench results as directional evidence for efficiency generalization, not as a full replication of the quality trade-offs in the main benchmark.

\subsection{Judge Sensitivity of Absolute Quality Scores}
\label{sec:judge_robustness}

Appendix Table~\ref{tab:judge_sensitivity} shows that absolute rubric-based
quality scores can vary substantially with judge choice, even for the same
generated reports. We therefore interpret the reported quality values primarily
as relative comparisons under a fixed independent judge, rather than as
judge-invariant absolute measurements. This caveat affects quality-sensitive
method comparisons more than the paper's efficiency conclusions, which do not
depend on rubric calibration. A stronger robustness analysis would test whether
method-level rankings remain stable across multiple independent judges on a
shared subset of examples; we leave that extension to future work.

\subsection{Sensitivity to Threshold Choice}
\label{sec:threshold_sensitivity}

To assess whether the reported operating points are unusually fragile, we perform a local threshold sweep for five representative post-retrieval pruning methods: \emph{MMR}, \emph{GRN}, \emph{Centroid Drift}, \emph{DPP}, and \emph{Submodular Coverage}. For each method, we fix all non-threshold hyperparameters at the published setting and vary only the pruning threshold over at least five values around the main-paper configuration. We report  the full sweep (Table~\ref{tab:threshold_sweep_points}).

We define a method's \emph{stable sampled operating region} as the largest contiguous threshold interval in the sampled sweep containing the published threshold $\tau_{\mathrm{pub}}$ for which overall quality remains within 2\% of the published configuration, i.e.,
\[
Q(\tau) \ge 0.98\,Q(\tau_{\mathrm{pub}}).
\]
This criterion tests whether the reported hyperparameters lie in a locally stable quality--efficiency region rather than at a brittle point estimate. Values below are mean scores over a 10-query sensitivity subset and are intended to characterize local stability rather than fully re-rank methods on the full benchmark.

%% file: sections/09_prompts.tex
\section{Prompts}
\label{sec:prompts}

\scriptsize

\begin{promptbox}
\textbf{You are an expert research-planning assistant.}

You will be provided with a root query, a current frame query, an optional parent research goal, and a set of candidate search queries with associated research goals. Your task is to judge which candidate queries are worth expanding in a deep research pipeline.

\textbf{Evaluation criteria:}
\begin{itemize}[leftmargin=1.2em,itemsep=1pt,topsep=2pt,parsep=0pt,partopsep=0pt]
    \item Prefer queries that are likely to retrieve high-value, non-redundant evidence.
    \item Favor candidates that expand topical coverage, clarify missing aspects of the problem, or are likely to lead to useful sources.
    \item Penalize candidates that are redundant, vague, off-topic, or unlikely to add meaningful new information.
\end{itemize}

\textbf{For each candidate, estimate:}
\begin{itemize}[leftmargin=1.2em,itemsep=1pt,topsep=2pt,parsep=0pt,partopsep=0pt]
    \item \texttt{predicted\_gain}: a score between 0 and 1 indicating expected marginal value.
    \item \texttt{keep\_probability}: a score between 0 and 1 indicating how strongly the candidate should be kept.
    \item \texttt{decision}: either \texttt{keep} or \texttt{prune}.
    \item \texttt{reason}: a short justification.
\end{itemize}

\textbf{Instructions:}
\begin{itemize}[leftmargin=1.2em,itemsep=1pt,topsep=2pt,parsep=0pt,partopsep=0pt]
    \item Judge usefulness semantically, not by surface wording alone.
    \item Be conservative: only keep queries that are likely to contribute meaningful new evidence.
    \item Return only a JSON object with no additional explanation.
\end{itemize}

\textbf{Output format:}

{\ttfamily\scriptsize
\{\\
\hspace*{1em}"candidates": [\\
\hspace*{2em}\{\\
\hspace*{3em}"idx": 1,\\
\hspace*{3em}"predicted\_gain": 0.81,\\
\hspace*{3em}"keep\_probability": 0.92,\\
\hspace*{3em}"decision": "keep",\\
\hspace*{3em}"reason": "high value"\\
\hspace*{2em}\}\\
\hspace*{1em}]\\
\}
}

\textbf{Inputs:}\\
Root query: \texttt{<root\_query>}\\
Current frame query: \texttt{<frame\_query>}\\
Parent research goal: \texttt{<parent\_research\_goal>}

\textbf{Candidate queries:}

{\ttfamily\scriptsize
1. Query: <candidate\_1\_query>\\
\hspace*{1em}Goal: <candidate\_1\_goal>\\[2pt]
2. Query: <candidate\_2\_query>\\
\hspace*{1em}Goal: <candidate\_2\_goal>\\[2pt]
...
}
\end{promptbox}

\smallskip
\begin{center}
\textit{(a) LLM-based query pruning instructions.}
\end{center}

\medskip

\begin{promptbox}
\textbf{You are an impartial evaluation assistant.}

You will be provided with a candidate question, a golden response (the ideal answer), and a candidate response (the model's answer to evaluate). Your task is to assign an alignment score indicating how well the candidate response matches the golden response.

\textbf{Scoring guidelines (0--5):}
\begin{itemize}[leftmargin=1.2em,itemsep=1pt,topsep=2pt,parsep=0pt,partopsep=0pt]
    \item \textbf{5} --- Fully aligned. Semantically equivalent; no missing key information; no contradictions.
    \item \textbf{4} --- Mostly aligned. Minor omissions or differences, but overall meaning preserved.
    \item \textbf{3} --- Partially aligned. Contains some correct elements but lacks important information.
    \item \textbf{2} --- Weak alignment. Only small portions match; the majority is incomplete or off-target.
    \item \textbf{1} --- Barely aligned. Very limited semantic overlap.
    \item \textbf{0} --- Not aligned. Irrelevant, contradictory, or does not follow the intent of the golden response (e.g., golden response asks for clarification, but the candidate provides a direct answer).
\end{itemize}

\textbf{Instructions:}
\begin{itemize}[leftmargin=1.2em,itemsep=1pt,topsep=2pt,parsep=0pt,partopsep=0pt]
    \item Judge semantic meaning, not surface wording.
    \item Be strict: only credit what the golden response explicitly or implicitly contains.
    \item Output only a single integer from 0 to 5 with no additional explanation.
\end{itemize}

\textbf{Inputs:}\\
Candidate question: \texttt{<question>}\\
Golden response: \texttt{<golden\_clarifying>}\\
Candidate response: \texttt{<candidate\_response>}
\end{promptbox}

\smallskip
\begin{center}
\textit{(b) LLM-as-a-judge evaluation instructions.}
\end{center}

\captionof{figure}{Prompts used in our pipeline.}
\label{fig:prompts}

\normalsize

%% file: sections/10_others.tex
\begin{table*}[t]
\centering
\caption{Judge-sensitivity analysis for rubric-based quality evaluation. Absolute scores vary substantially with judge choice, so main-paper quality results should be interpreted as relative comparisons under a fixed judge rather than as judge-invariant quality values.}

\label{tab:judge_sensitivity}
\small
\setlength{\tabcolsep}{5pt}
\renewcommand{\arraystretch}{1.15}
\resizebox{\textwidth}{!}{
\begin{tabular}{l c ccccccc c c}
\toprule
\textbf{Model} & \textbf{Judge} & \textbf{Overall} & \textbf{Clarity} & \textbf{Depth} & \textbf{Balance} & \textbf{Breadth} & \textbf{Support} & \textbf{Insight} & \textbf{KPR + KPC} & \textbf{Cit. Recall} \\
\midrule

gpt-5-mini & gpt-5-mini & 58.47 & 57.20 & 61.50 & 59.90 & 67.40 & 41.50 & 63.30 & 61.70 & 92.98 \\

gpt-4.1-mini & gpt-4.1-mini & 91.50 & 89.00 & 96.00 & 90.00 & 96.60 & 91.00 & 87.00 & 72.24 & 99.52 \\
gpt-4.1-mini & gpt-5-mini & 47.67 & 50.00 & 45.00 & 48.00 & 61.00 & 32.00 & 50.00 & 72.25 & 90.56 \\

Qwen2-7B-Instruct (32k) & gpt-5-mini & 34.37 & 35.71 & 35.24 & 39.05 & 42.86 & 21.43 & 31.90 & 72.55 & 75.45 \\

\bottomrule
\end{tabular}
}
\end{table*}

\begin{table*}[t]
\centering
\caption{Performance comparison across all pruning strategies. Tokens are reported in thousands (k) (\S~\ref{sec:token_cost}). Values are mean ± standard error over 100 reports (\S~\ref{sec:metric_computation}). Quality, relevance, and faithfulness metrics were obtained from DeepResearchGym (\S~\ref{sec:qual_rel_faith})}
\label{tab:quality_all}
\small
\setlength{\tabcolsep}{5pt}
\renewcommand{\arraystretch}{1.15}
\resizebox{\textwidth}{!}{
\begin{tabular}{l cc ccccccc c c}
\toprule
& \multicolumn{2}{c}{\textbf{Cost}}
& \multicolumn{7}{c}{\textbf{Quality}}
& \textbf{Relevance}
& \textbf{Faithfulness} \\
\cmidrule(lr){2-3} \cmidrule(lr){4-10} \cmidrule(lr){11-11} \cmidrule(lr){12-12}
\textbf{Method}
& \# Tokens & Runtime (s) & Overall & Clarity & Depth & Balance & Breadth & Support & Insight & KPR + KPC & Cit. Recall\\
\midrule

Baseline & 375.4 $\pm$ 2.6k & 3422.6 $\pm$ 140.6 & 57.83 & 55.05 & 62.12 & 57.27 & 65.96 & 45.45 & 61.11 & \textbf{70.23} & 95.54 \\

\midrule
\multicolumn{12}{l}{\textbf{One-stage methods: Post-Retrieval Pruning}} \\
MMR(\S~\ref{sec:mmr}) & 114.6 $\pm$ 2.2k & 1379.8 $\pm$ 68.3 & 56.62 & 54.50 & 59.90 & 57.20 & 65.50 & 43.20 & 59.40 & 63.49 & 91.70 \\
Geo.\ Residual Novelty(\S~\ref{sec:geometric_residual_novelty}) & 175.7 $\pm$ 5.1k & 2087.9 $\pm$ 108.3 & 57.02 & 55.85 & 59.36 & 58.30 & 63.72 & 44.15 & 60.74 & 42.89 & 94.40 \\

Centroid Drift(\S~\ref{sec:centroid_drift})  & 137.5 $\pm$ 2.0k & 1605.7 $\pm$ 73.9 & 56.84 & 56.12 & 59.90 & 56.63 & 64.18 & 43.37 & 60.82 & 41.29 & 92.48 \\

Submodular Coverage(\S~\ref{sec:submodular_coverage}) & 141.9 $\pm$ 3.3k & 1737.2 $\pm$ 108.1 & 55.22 & 53.54 & 57.68 & 55.86 & 62.42 & 42.63 & 59.19 & 47.94 & 91.97 \\

DPP(\S~\ref{sec:dpp})  & 129.6 $\pm$ 2.5k & 1520.6 $\pm$ 73.3 & 54.51 & 53.19 & 56.81 & 55.71 & 62.75 & 41.10 & 57.47 & 43.33 & 95.62 \\
Combined (\S~\ref{sec:combined_scoring})& 117.8 $\pm$ 2.5k & 1399.5 $\pm$ 71.0& 56.03 & 54.90 & 58.30 & 57.20 & 64.20 & 42.60 & 59.00 & 64.13 & 92.59 \\
Hybrid(\S~\ref{sec:combined_scoring}) & 121.3 $\pm$ 2.8k & 1449.5 $\pm$ 75.6  & 57.63 & 57.40 & 59.70 & 57.90 & 65.10 & 45.10 & 60.60 & 65.23 & 92.16 \\
LLM (\S~\ref{sec:lexical_pruning}) & 211.8 $\pm$ 7.9k & 2310.7 $\pm$ 137.8 & 59.65 & 57.40 & 63.00 & 60.00 & 66.80 & 48.20 & 62.50 & 49.65 & 93.54 \\

\midrule
\multicolumn{12}{l}{\textbf{One-stage methods: Pre-Synthesis Pruning}} \\
MMR & 366.0 $\pm$ 5.2k & 4446.9 $\pm$ 159.0 & 57.77 & 55.30 & 59.20 & 60.00 & 66.70 & 44.50 & 60.90 & 65.26 & 90.41 \\
Geo.\ Residual Novelty & 374.1 $\pm$ 3.1k & 4493.3 $\pm$ 155.2 & 55.98 & 53.70 & 59.35 & 56.41 & 64.67 & 41.96 & 59.78 & 41.67 & 90.38 \\

Centroid Drift & 374.5 $\pm$ 2.8k & 4562.4 $\pm$ 149.8 & 52.38 & 49.60 & 55.70 & 52.60 & 60.50 & 40.60 & 55.30 & 43.75 & 90.19 \\

Submodular Coverage & 384.7 $\pm$ 3.3k & 4533.3 $\pm$ 150.5 & 53.22 & 50.40 & 55.25 & 53.64 & 62.12 & 40.91 & 56.97 & 42.09 & 90.78 \\

DPP & 374.4 $\pm$ 3.2k & 4512.4 $\pm$ 149.8 & 54.70 & 52.12 & 57.37 & 56.26 & 63.64 & 41.52 & 57.27 & 54.07 & 90.24 \\
Combined & 374.5 $\pm$ 6.1k & 4408.6 $\pm$ 161.6  & 59.38 & 56.70 & 62.40 & 59.90 & 68.10 & 46.60 & 62.60 & 66.32 & 93.79 \\
Hybrid & 332.3 $\pm$ 10.6k & 3834.1 $\pm$ 205.6 & 60.68 & 57.50 & 64.10 & 61.80 & 70.00 & 46.40 & 64.30 & 65.62 & 95.07 \\
LLM & 386.7 $\pm$ 3.1k & 4512.0 $\pm$ 153.7 & 57.17 & 54.50 & 60.00 & 57.90 & 65.40 & 44.50 & 60.70 & 44.47 & 92.43 \\

\midrule
\multicolumn{12}{l}{\textbf{Two-stage methods: Post-Retrieval + Pre-Synthesis Pruning}} \\

MMR & 114.6 $\pm$ 2.2k & 1381.3 $\pm$ 68.2 & 56.40 & 55.90 & 58.80 & 57.20 & 64.70 & 43.70 & 58.10 & 65.16 & 92.61 \\
Geo.\ Residual Novelty  & 175.2 $\pm$ 5.3k & 2080.2 $\pm$ 107.5 & 57.94 & 54.37 & 60.52 & 58.33 & 64.48 & 43.02 & 60.94 & 64.47 & 93.06 \\


Centroid Drift  & 118.1 $\pm$ 1.3k & 1142.43 $\pm$ 105.82 & 57.44 & 55.57 & 59.48 & 59.69 & 64.95 & 44.54 & 60.41 & 43.23 & 92.94 \\

Submodular Coverage  & 142.1 $\pm$ 3.2k & 1776.5 $\pm$ 112.1 & 57.00 & 55.50 & 59.40 & 57.70 & 65.20 & 43.80 & 60.40 & 62.23 & 94.74 \\

DPP  & 122.8 ± 1.8k & 1462.31 ± 236.55 & 56.73 & 56.60 & 58.40 & 56.90 & 64.60 & 44.50 & 59.40 & 62.33 & 93.62 \\

CD + SC & 137.5 $\pm$ 2.0k & 1599.6 $\pm$ 73.6 & 59.47 & 58.70 & 57.80 & 56.80 & 63.60 & 40.20 & 59.70 & 44.29 & 89.96 \\

CD + LLM & 136.0 $\pm$ 3.2k & 1589.3 $\pm$ 76.2 & 58.65 & 56.80 & 62.10 & 58.90 & 66.20 & 46.00 & 61.90 & 63.40 & 94.34 \\
SC + LLM & 139.1 $\pm$ 4.1k & 1700.6 $\pm$ 110.0 & 58.08 & 55.90 & 61.50 & 58.60 & 65.80 & 45.50 & 61.20 & 64.54 & 93.43 \\
LLM  & 226.6 $\pm$ 8.1k & 2632.3 $\pm$ 148.2 & 58.27 & 55.05 & 61.11 & 58.89 & 66.26 & 46.06 & 62.22 & 48.61 & 93.01 \\

Lexical & 115.1 $\pm$ 1.2k & 1414.3 $\pm$ 64.8 & 54.62 & 53.40 & 57.20 & 57.30 & 61.70 & 40.30 & 57.80 & 46.13 & 91.38 \\

Combined & 117.6 $\pm$ 2.6k & 1392.2 $\pm$ 71.4 & 54.87 & 53.90 & 57.50 & 56.00 & 64.20 & 39.80 & 57.80 & 64.88 & 93.03 \\
Hybrid & 121.7 $\pm$ 2.8k & 1457.4 $\pm$ 75.3 & 56.92 & 56.30 & 60.20 & 57.30 & 65.40 & 43.20 & 59.10 & 64.25 & 93.51\\

\midrule
\multicolumn{12}{l}{\textbf{Three-stage methods: Pre-Retrieval + Post-Retrieval + Pre-Synthesis Pruning}} \\
MMR & 100.1 $\pm$ 2.0k & 1157.7 $\pm$ 63.6 & 55.90 & 54.00 & 58.90 & 56.30 & 63.70 & 42.60 & 59.90 & 63.43 & 91.84 \\
Geo.\ Residual Novelty & 150.0 $\pm$ 4.5k & 1937.7 $\pm$ 344.0 & 56.68 & 54.55 & 58.69 & 58.18 & 64.85 & 43.64 & 60.20 & 45.13 & 91.33 \\

Centroid Drift & 120.6 $\pm$ 2.1k & 1669.8 $\pm$ 321.4 & 56.00 & 54.90 & 58.98 & 55.92 & 63.78 & 42.24 & 60.20 & 43.64 & 93.30 \\

Submodular & 121.8 $\pm$ 2.8k & 1438.9 $\pm$ 106.1 & 55.52 & 53.23 & 58.79 & 56.57 & 64.14 & 41.52 & 58.89 & 65.79 & 92.03 \\
DPP & 113.3 $\pm$ 2.2k & 1626.2 $\pm$ 356.3 & 55.41 & 54.08 & 57.76 & 56.33 & 62.14 & 43.88 & 58.27 & 46.33 & 95.40 \\
Lexical + CD + SC & 120.1 $\pm$ 2.3k & 1375.0 $\pm$ 63.4 & 57.81 & 57.17 & 59.49 & 59.09 & 64.85 & 45.05 & 61.21 & 42.67 & 93.31 \\
Combined & 102.8 $\pm$ 2.2k & 1227.6 $\pm$ 76.6 & 55.00 & 54.00 & 57.50 & 57.80 & 63.80 & 39.20 & 57.70 & 62.50 & 90.83 \\
Hybrid & 106.3 $\pm$ 2.6k & 1282.7 $\pm$ 81.2  & 57.10 & 56.60 & 59.90 & 57.10 & 64.90 & 44.50 & 59.60 & 64.80 & 93.06 \\
Learned Query + Geo.\ Residual Novelty & 145.7 $\pm$ 4.3k & 1744.4 $\pm$ 103.5 & 58.13 & 56.60 & 61.30 & 59.00 & 65.60 & 45.80 & 60.50 & 51.78 & 95.48 \\
LLM & 143.6 $\pm$ 7.8k & $1575.2 \pm 120.2$ & 59.53 & 58.17 & 62.37 & 60.43 & 66.56 & 46.77 & 62.90 & 65.09 & 93.12 \\

Lexical & 99.0 $\pm$ 1.1k & 1166.4 $\pm$ 56.7 & 54.82 & 54.80 & 57.50 & 56.30 & 62.10 & 40.40 & 57.80 & 46.56 & 93.10 \\

\bottomrule
\end{tabular}}

\vspace{4pt}

\begin{minipage}{0.95\textwidth}
\footnotesize
Unless otherwise specified, the same pruning criterion is applied at both the post-retrieval and pre-synthesis stages. Method-specific hyperparameters are: MMR ($\lambda=0.35$, max 10 contexts); DPP ($\tau=0.30$, max 10 pre-synthesis contexts); centroid drift ($\delta=0.03$); geometric residual novelty (GRN, $\tau=0.85$); and submodular coverage (prune when marginal gain per token $< 0.05$);LLM ($\delta=0.3$); Lexical ($\lambda=.6$, threshold = .2).
\end{minipage}

\end{table*}

\vspace{10pt}
\begin{table*}[t]
\centering
\caption{Efficiency comparison across all methods. Tokens are reported in thousands (k). Est. Token Reduction (\%) denotes the estimated reduction relative to each method's pre-pruning budget. Savings vs. Baseline (\%) is computed from mean total tokens relative to the baseline mean token count.}
\label{tab:efficiency_methods_10}
\small
\setlength{\tabcolsep}{5pt}
\renewcommand{\arraystretch}{1.15}
\resizebox{\textwidth}{!}{
\begin{tabular}{l cccc}
\toprule

\textbf{Method} & \textbf{\# Nodes} (\S~\ref{sec:num_nodes}) & \textbf{\# Tokens} (\S~\ref{sec:token_cost}) & \textbf{Est. Token Reduction (\%)}(\S~\ref{sec:est_token_reduction})& \textbf{Savings vs. Baseline (\%)}(\S~\ref{sec:est_save_baseline}) \\
\midrule

Baseline & 29.0 $\pm$ 0.0 & 375.4 $\pm$ 2.6k & 0.0 & 0.0 \\
\midrule
\multicolumn{5}{l}{\textbf{One-stage methods: Post-Retrieval Pruning}} \\
Geo.\ Residual Novelty (\S~\ref{sec:geometric_residual_novelty}) & 13.31 $\pm$ 0.37 & 175.7 $\pm$ 5.1k & 22.02 $\pm$ 0.26 & 53.2 \\
Centroid Drift (\S~\ref{sec:centroid_drift}) & 10.47 $\pm$ 0.13 & 137.5 $\pm$ 2.0k & 18.91 $\pm$ 0.26 & 63.4 \\
Submodular Coverage (\S~\ref{sec:submodular_coverage})& 10.82 $\pm$ 0.28 & 141.9 $\pm$ 3.3k & 21.55 $\pm$ 0.26 & 62.2 \\
DPP (\S~\ref{sec:dpp})& 9.88 $\pm$ 0.16 & 129.6 $\pm$ 2.5k & 21.99 $\pm$ 0.22 & 65.5 \\
LLM (\S~\ref{sec:llm_query_pruning})& 14.90 $\pm$ 0.53 & 211.8 $\pm$ 7.9k & 22.76 $\pm$ 0.38 & 43.6 \\
MMR(\S~\ref{sec:mmr}) & 8.84 $\pm$ 0.08 & 114.6 $\pm$ 2.2k & 22.11 $\pm$ 0.48 & 69.5 \\
Combined (\S~\ref{sec:combined_scoring})& 9.00 $\pm$ 0.11 & 117.8 $\pm$ 2.5k & 21.01 $\pm$ 0.48 & 68.6 \\
Hybrid(\S~\ref{sec:combined_scoring}) & 9.28 $\pm$ 0.15 & 121.3 $\pm$ 2.8k & 20.92 $\pm$ 0.49 & 67.7 \\

\midrule
\multicolumn{5}{l}{\textbf{One-stage methods: Pre-Synthesis Pruning}} \\
Geo.\ Residual Novelty & 29 $\pm$ 0.00 & 374.1 $\pm$ 3.1k & 28.12 $\pm$ 0.17 & 0.4 \\
Centroid Drift & 29 $\pm$ 0.00 & 374.5 $\pm$ 2.8k & 28.12 $\pm$ 0.17 & 0.2 \\
Submodular Coverage & 29 $\pm$ 0.00 & 384.7 $\pm$ 3.3k & 27.58 $\pm$ 0.18 & -2.5 \\
DPP & 29 $\pm$ 0.00 & 374.4 $\pm$ 3.2k & 28.15 $\pm$ 0.18 & 0.3 \\
LLM & 29 $\pm$ 0.00 & 386.7 $\pm$ 3.1k & 27.66 $\pm$ 0.25 & -3.0 \\
MMR & 29 $\pm$ 0.00 & 366.0 $\pm$ 5.2k & 27.17 $\pm$ 0.55 & 2.5 \\
Hybrid & 29 $\pm$ 0.00 & 332.3 $\pm$ 10.6k & 24.62 $\pm$ 1.26 & 11.5 \\
Combined & 29 $\pm$ 0.00 & 374.5 $\pm$ 6.1k & 30.31 $\pm$ 0.66 & 0.2 \\

\midrule
\multicolumn{5}{l}{\textbf{Two-stage methods: Post-Retrieval + Pre-Synthesis Pruning}} \\
Centroid Drift & 9.02 $\pm$ 0.02 & 118.1 $\pm$ 1.3k & 20.58 $\pm$ 0.24 & 68.5 \\
Submodular Coverage & 10.80 $\pm$ 0.28 & 142.1 $\pm$ 3.2k & 22.92 $\pm$ 0.29 & 62.1 \\
CD + SC & 10.45 $\pm$ 0.13 & 137.5 $\pm$ 2.0k & 20.69 $\pm$ 0.33 & 63.4 \\
LLM & 15.57 $\pm$ 0.53 & 226.6 $\pm$ 8.1k & 23.57 $\pm$ 0.31 & 39.7 \\
Lexical & 9.00 $\pm$ 0.00 & 115.1 $\pm$ 1.2k & 29.36 $\pm$ 0.24 & 69.3 \\
MMR & 8.84 $\pm$ 0.08 & 114.6 $\pm$ 2.2k & 22.11 $\pm$ 0.48 & 69.5 \\
Hybrid & 9.30 $\pm$ 0.15 & 121.7 $\pm$ 2.8k & 26.31 $\pm$ 0.66 & 67.6 \\
Combined & 8.98 $\pm$ 0.11 & 117.6 $\pm$ 2.6k & 27.81 $\pm$ 0.60 & 68.7 \\
CD + LLM & 10.24 $\pm$ 0.17 & 136.0 $\pm$ 3.2k & 20.30 $\pm$ 0.51 & 63.8 \\
SC + LLM & 10.54 $\pm$ 0.30 & 139.1 $\pm$ 4.1k & 22.40 $\pm$ 0.55 & 62.9 \\

\midrule
\multicolumn{5}{l}{\textbf{Three-stage methods: Pre-Retrieval + Post-Retrieval + Pre-Synthesis Pruning}} \\
Geo.\ Residual Novelty  & 11.57 $\pm$ 0.33 & 150.0 $\pm$ 4.5k & 19.97 $\pm$ 0.33 & 60.0 \\
DPP  & 8.68 $\pm$ 0.15 & 113.3 $\pm$ 2.2k & 20.16 $\pm$ 0.23 & 69.8 \\
Centroid Drift & 9.27 $\pm$ 0.13 & 120.6 $\pm$ 2.1k & 16.65 $\pm$ 0.28 & 67.9 \\
Lexical + CD + SC & 9.26 $\pm$ 0.14 & 120.1 $\pm$ 2.3k & 17.82 $\pm$ 0.39 & 68.0 \\
Lexical & 7.90 $\pm$ 0.04 & 99.0 $\pm$ 1.1k & 28.72 $\pm$ 0.27 & 73.6 \\
MMR  & 7.82 $\pm$ 0.08 & 100.1 $\pm$ 2.0k & 20.43 $\pm$ 0.44 & 73.3 \\
Hybrid & 8.16 $\pm$ 0.14 & 106.3 $\pm$ 2.6k & 25.56 $\pm$ 0.66 & 71.7 \\
Combined & 7.90 $\pm$ 0.10 & 102.8 $\pm$ 2.2k & 27.33 $\pm$ 0.58 & 72.6 \\
LLM & 10.12 $\pm$ 0.45 & 143.9 $\pm$ 7.7k & 19.45 $\pm$ 0.81 & 61.7 \\

\bottomrule
\end{tabular}
}
\end{table*}

\begin{table*}[t]
\centering
\caption{Runtime breakdown across all methods.(\S~\ref{sec:runtime_metrics}) Values are reported as mean $\pm$ standard error over 100 reports (\S~\ref{sec:metric_computation}). Research / Scraping denotes the retrieval-heavy stage (`branch\_research` for the baseline and `scraping` for pruning methods). Pre-Retrieval, Post-Retrieval, and Pre-Synthesis report the latency of the corresponding pruning stages when separately instrumented.}
\label{tab:runtime_breakdown_metric}
\small
\setlength{\tabcolsep}{5pt}
\renewcommand{\arraystretch}{1.15}
\resizebox{\textwidth}{!}{
\begin{tabular}{l cccccccc}
\toprule
\textbf{Method} & \textbf{Total Runtime (s)} & \textbf{Research / Scraping} & \textbf{Result Proc.} & \textbf{Query Gen} & \textbf{Planning} & \textbf{Pre-Retrieval} & \textbf{Post-Retrieval} & \textbf{Pre-Synthesis} \\
\midrule

Baseline & 3422.6 $\pm$ 140.6 & 2888.2 $\pm$ 304.2 & 1158.5 $\pm$ 54.3 & 339.0 $\pm$ 13.3 & 13.3 $\pm$ 0.7 & -- & -- & -- \\
\midrule
\multicolumn{9}{l}{\textbf{One-stage methods: Post-Retrieval Pruning}} \\
Geo.\ Residual Novelty & 1889.05 $\pm$ 400.04 & 1321.83 $\pm$ 317.06 & 456.61 $\pm$ 58.99 & 122.68 $\pm$ 18.15 & 13.47 $\pm$ 0.65 & -- & 2.98 $\pm$ 0.33 & -- \\
Centroid Drift & 1587.48 $\pm$ 220.50 & 1098.06 $\pm$ 224.15 & 377.13 $\pm$ 23.32 & 92.22 $\pm$ 5.42 & 13.62 $\pm$ 0.69 & -- & 2.46 $\pm$ 0.09 & -- \\
Submodular Coverage & 1734.63 $\pm$ 246.68 & 1209.74 $\pm$ 235.58 & 413.96 $\pm$ 34.39 & 105.59 $\pm$ 12.63 & 13.62 $\pm$ 0.69 & -- & 3.03 $\pm$ 0.39 & -- \\
DPP & 1242.07 $\pm$ 152.57 & 823.06 $\pm$ 114.47 & 340.68 $\pm$ 27.93 & 80.77 $\pm$ 7.58 & 13.48 $\pm$ 0.66 & -- & 2.59 $\pm$ 0.15 & -- \\
LLM & 2469.87 $\pm$ 446.85 & 1499.30 $\pm$ 333.19 & 671.39 $\pm$ 95.77 & 187.37 $\pm$ 30.62 & 13.62 $\pm$ 0.69 & 105.64 $\pm$ 18.95 & -- & -- \\
Combined & $1399.50 \pm 71.00$ & $999.09 \pm 64.98$ & $310.68 \pm 8.18$ & $75.29 \pm 2.48$ & $14.03 \pm 0.47$ & -- & $1.63 \pm 0.06$ & -- \\
Hybrid & $1449.50 \pm 75.61$ & $1038.10 \pm 68.81$ & $318.96 \pm 8.31$ & $78.33 \pm 2.69$ & $14.03 \pm 0.47$ & -- & $2.50 \pm 0.10$ & -- \\
MMR & $1121.78 \pm 136.98$ & $740.04 \pm 106.95$ & $304.02 \pm 28.54$ & $69.35 \pm 7.24$ & $12.92 \pm 0.79$ & -- & $2.45 \pm 0.09$ & -- \\
\midrule
\multicolumn{9}{l}{\textbf{One-stage methods: Pre-Synthesis Pruning}} \\
Geo.\ Residual Novelty & 4216.31 $\pm$ 473.91 & 2796.29 $\pm$ 359.11 & 1111.71 $\pm$ 88.83 & 330.73 $\pm$ 17.85 & 13.43 $\pm$ 0.68 & -- & -- & 6.34 $\pm$ 3.87 \\
Centroid Drift & 4406.07 $\pm$ 334.44 & 2888.23 $\pm$ 304.19 & 1158.54 $\pm$ 54.34 & 339.04 $\pm$ 13.29 & 13.27 $\pm$ 0.70 & -- & -- & 2.27 $\pm$ 0.06 \\
Submodular Coverage & 4370.96 $\pm$ 354.75 & 2904.30 $\pm$ 296.47 & 1165.94 $\pm$ 50.74 & 343.32 $\pm$ 12.54 & 13.62 $\pm$ 0.69 & -- & -- & 1.29 $\pm$ 0.12 \\
DPP & 4365.29 $\pm$ 358.83 & 2896.47 $\pm$ 300.15 & 1161.14 $\pm$ 52.98 & 342.91 $\pm$ 12.55 & 13.62 $\pm$ 0.69 & -- & -- & 2.23 $\pm$ 0.15 \\
LLM & 4345.87 $\pm$ 384.21 & 2862.36 $\pm$ 317.91 & 1142.62 $\pm$ 64.34 & 333.12 $\pm$ 16.24 & 13.62 $\pm$ 0.69 & -- & -- & 20.23 $\pm$ 2.65 \\
Combined & 4408.62 $\pm$ 161.63 & 3002.82 $\pm$ 143.58 & 1091.47 $\pm$ 22.32 & 322.51 $\pm$ 6.69 & 14.00 $\pm$ 0.47 & -- & -- & 1.26 $\pm$ 0.06 \\
Hybrid & 3834.09 $\pm$ 205.60 & 2625.22 $\pm$ 164.30 & 999.33 $\pm$ 30.68 & 303.32 $\pm$ 8.32 & 13.37 $\pm$ 0.50 & -- & -- & 2.38 $\pm$ 0.09 \\
MMR & 4446.91 $\pm$ 158.95 & 3028.76 $\pm$ 142.72 & 1096.20 $\pm$ 21.75 & 323.49 $\pm$ 6.65 & 14.07 $\pm$ 0.47 & -- & -- & 2.52 $\pm$ 0.06 \\
\midrule
\multicolumn{9}{l}{\textbf{Two-stage methods: Post-Retrieval + Pre-Synthesis Pruning}} \\
Centroid Drift  & 1142.43 $\pm$ 105.82 & 748.48 $\pm$ 88.49 & 335.49 $\pm$ 14.29 & 77.85 $\pm$ 5.58 & 14.43 $\pm$ 1.25 & -- & 1.33 $\pm$ 0.15 & 0.83 $\pm$ 0.03 \\
Submodular Coverage & 1513.36 $\pm$ 163.23 & 1020.51 $\pm$ 146.46 & 426.78 $\pm$ 37.31 & 107.07 $\pm$ 14.48 & 13.87 $\pm$ 0.83 & -- & 1.42 $\pm$ 0.15 & 0.20 $\pm$ 0.05 \\
Centroid Drift + Submodular & 1588.01 $\pm$ 220.44 & 1098.06 $\pm$ 224.15 & 377.13 $\pm$ 23.32 & 92.22 $\pm$ 5.42 & 13.62 $\pm$ 0.69 & -- & 2.57 $\pm$ 0.17 & 0.29 $\pm$ 0.01 \\
LLM & 2962.11 $\pm$ 434.18 & 1763.21 $\pm$ 337.56 & 769.32 $\pm$ 85.00 & 210.68 $\pm$ 28.73 & 13.60 $\pm$ 0.69 & -- & 177.24 $\pm$ 21.10 & 24.07 $\pm$ 1.84 \\
Lexical & 1414.3 $\pm$ 64.8 & 1014.8 $\pm$ 64.2 & 313.1 $\pm$ 5.8 & 76.0 $\pm$ 1.8 & 14.2 $\pm$ 0.4 & -- & 0.04 $\pm$ 0.00 & 0.01 $\pm$ 0.00 \\
SC+LLM & $1700.60 \pm 109.98$ & $1210.75 \pm 97.39$ & $368.65 \pm 13.67$ & $95.10 \pm 4.44$ & $13.97 \pm 0.47$ & -- & $3.99 \pm 0.11$ & $9.61 \pm 0.33$ \\
CD+LLM & $1589.33 \pm 76.18$ & $1109.23 \pm 68.27$ & $362.02 \pm 10.28$ & $90.84 \pm 3.05$ & $13.97 \pm 0.47$ & -- & $3.65 \pm 0.07$ & $10.80 \pm 0.24$ \\
Combined & $1392.21 \pm 71.36$ & $994.38 \pm 65.22$ & $309.02 \pm 8.44$ & $74.78 \pm 2.45$ & $14.04 \pm 0.46$ & -- & $1.60 \pm 0.06$ & $0.27 \pm 0.01$ \\
Hybrid & $1457.36 \pm 75.31$ & $1043.03 \pm 68.60$ & $320.21 \pm 8.36$ & $78.93 \pm 2.69$ & $14.03 \pm 0.47$ & -- & $2.34 \pm 0.08$ & $0.35 \pm 0.01$ \\
MMR & $1381.30 \pm 68.22$ & $987.79 \pm 63.59$ & $305.04 \pm 6.69$ & $73.10 \pm 2.15$ & $14.09 \pm 0.46$ & -- & $2.45 \pm 0.03$ & $1.00 \pm 0.03$ \\

\midrule
\multicolumn{9}{l}{\textbf{Three-stage methods: Pre-Retrieval + Post-Retrieval + Pre-Synthesis Pruning}} \\
Geo.\ Residual Novelty & 1937.70 $\pm$ 344.04 & 1417.29 $\pm$ 318.62 & 384.96 $\pm$ 40.14 & 112.58 $\pm$ 14.24 & 13.62 $\pm$ 0.69 & 0.85 $\pm$ 0.07 & 2.62 $\pm$ 0.30 & 1.33 $\pm$ 0.07 \\
DPP & 1626.15 $\pm$ 356.25 & 1189.83 $\pm$ 333.62 & 326.32 $\pm$ 29.25 & 88.16 $\pm$ 6.03 & 13.62 $\pm$ 0.69 & 0.78 $\pm$ 0.05 & 2.22 $\pm$ 0.21 & 1.24 $\pm$ 0.19 \\
Centroid Drift & 1669.84 $\pm$ 321.43 & 1212.22 $\pm$ 317.93 & 339.74 $\pm$ 22.14 & 95.95 $\pm$ 5.31 & 13.62 $\pm$ 0.69 & 0.89 $\pm$ 0.11 & 2.07 $\pm$ 0.16 & 1.36 $\pm$ 0.15 \\
Lexical + CD + SC & 1334.22 $\pm$ 294.81 & 908.79 $\pm$ 279.25 & 329.63 $\pm$ 24.07 & 91.36 $\pm$ 6.27 & 13.62 $\pm$ 0.69 & 0.00 $\pm$ 0.00 & 2.19 $\pm$ 0.06 & 0.29 $\pm$ 0.01 \\
Lexical & 1166.4 $\pm$ 56.7 & 806.3 $\pm$ 54.0 & 273.6 $\pm$ 5.8 & 74.6 $\pm$ 1.7 & 14.1 $\pm$ 0.4 & 0.00 $\pm$ 0.00 & 0.03 $\pm$ 0.00 & 0.01 $\pm$ 0.00 \\
LLM & $1575.23 \pm 120.22$ & $1038.86 \pm 95.46$ & $353.69 \pm 21.13$ & $99.04 \pm 6.47$ & $13.45 \pm 0.50$ & $23.62 \pm 1.22$ & $51.99 \pm 2.61$ & $9.11 \pm 0.43$ \\
Combined & $1227.60 \pm 76.62$ & $866.04 \pm 73.06$ & $271.15 \pm 7.03$ & $73.85 \pm 2.24$ & $14.11 \pm 0.47$ & $0.62 \pm 0.03$ & $1.30 \pm 0.03$ & $0.26 \pm 0.01$ \\
Hybrid & $1282.72 \pm 81.18$ & $910.23 \pm 76.40$ & $278.68 \pm 8.20$ & $77.03 \pm 2.37$ & $14.01 \pm 0.47$ & $1.06 \pm 0.04$ & $1.72 \pm 0.06$ & $0.35 \pm 0.01$ \\
MMR & $1157.70 \pm 63.60$ & $799.70 \pm 59.94$ & $267.54 \pm 6.88$ & $72.83 \pm 2.11$ & $14.10 \pm 0.47$ & $0.79 \pm 0.02$ & $2.03 \pm 0.03$ & $1.00 \pm 0.03$ \\
\bottomrule
\end{tabular}
}
\end{table*}

\begin{table*}[t]
\centering
\caption{Token breakdown across all methods (in thousands, k) (\S~\ref{sec:token_break_stage}). Values are mean ± standard error (\S~\ref{sec:metric_computation}).}
\label{tab:token_breakdown_10}
\small
\setlength{\tabcolsep}{5pt}
\renewcommand{\arraystretch}{1.15}
\resizebox{\textwidth}{!}{
\begin{tabular}{l cccccc}
\toprule
\textbf{Method} & \textbf{Total Tokens} & \textbf{Planning} & \textbf{Query Gen} & \textbf{Query Pruning} & \textbf{Result Proc.} & \textbf{Embedding} \\
\midrule

Baseline & 375.4 $\pm$ 2.6k & 2.3 $\pm$ 0.1k & 11.9 $\pm$ 0.1k & -- & 361.3 $\pm$ 2.5k & -- \\
\midrule
\multicolumn{7}{l}{\textbf{One-stage methods: Post-Retrieval Pruning}} \\
Geo.\ Residual Novelty & 175.7 $\pm$ 5.1k & 2.3 $\pm$ 0.1k & 4.8 $\pm$ 0.2k & -- & 163.2 $\pm$ 3.8k & 5.5 $\pm$ 0.2k \\
Centroid Drift & 137.5 $\pm$ 2.0k & 2.3 $\pm$ 0.1k & 3.5 $\pm$ 0.1k & -- & 127.5 $\pm$ 1.6k & 4.2 $\pm$ 0.1k \\
Submodular Coverage & 141.9 $\pm$ 3.3k & 2.3 $\pm$ 0.1k & 3.6 $\pm$ 0.1k & -- & 130.5 $\pm$ 2.4k & 5.5 $\pm$ 0.2k \\
DPP & 129.6 $\pm$ 2.5k & 2.3 $\pm$ 0.1k & 3.2 $\pm$ 0.1k & -- & 119.3 $\pm$ 1.8k & 4.9 $\pm$ 0.1k \\
LLM & 211.8 $\pm$ 7.9k & 2.3 $\pm$ 0.1k & 5.5 $\pm$ 0.3k & 20.8 $\pm$ 0.9k & 183.3 $\pm$ 5.4k & -- \\
Combined & 117.8 $\pm$ 2.5k & 2.2 $\pm$ 0.1k & 2.8 $\pm$ 0.1k & -- & 75.1 $\pm$ 1.7k & 5.7 $\pm$ 0.1k \\
Hybrid & 121.3 $\pm$ 2.8k & 2.2 $\pm$ 0.1k & 2.9 $\pm$ 0.1k & -- & 77.5 $\pm$ 1.9k & 5.9 $\pm$ 0.2k \\
MMR & 114.6 $\pm$ 2.2k & 2.2 $\pm$ 0.1k & 2.7 $\pm$ 0.0k & -- & 74.0 $\pm$ 1.6k & 4.3 $\pm$ 0.1k \\

\midrule
\multicolumn{7}{l}{\textbf{One-stage methods: Pre-Synthesis Pruning}} \\
Geo.\ Residual Novelty & 374.1 $\pm$ 3.1k & 2.2 $\pm$ 0.1k & 11.9 $\pm$ 0.1k & -- & 360.0 $\pm$ 2.7k & -- \\
Centroid Drift & 374.5 $\pm$ 2.8k & 2.3 $\pm$ 0.1k & 11.9 $\pm$ 0.1k & -- & 360.4 $\pm$ 2.6k & -- \\
Submodular Coverage & 384.7 $\pm$ 3.3k & 2.3 $\pm$ 0.1k & 11.9 $\pm$ 0.1k & -- & 358.7 $\pm$ 2.8k & 11.8 $\pm$ 0.1k \\
DPP & 374.4 $\pm$ 3.2k & 2.3 $\pm$ 0.1k & 11.9 $\pm$ 0.1k & -- & 360.3 $\pm$ 3.0k & -- \\
LLM & 386.7 $\pm$ 3.1k & 2.3 $\pm$ 0.1k & 11.9 $\pm$ 0.1k & 13.2 $\pm$ 0.2k & 359.3 $\pm$ 2.8k & -- \\
Combined & 374.5 $\pm$ 6.1k & 2.2 $\pm$ 0.1k & 11.8 $\pm$ 0.1k & -- & 248.4 $\pm$ 5.1k & -- \\
Hybrid & 332.3 $\pm$ 10.6k & 1.9 $\pm$ 0.1k & 11.6 $\pm$ 0.1k & -- & 211.4 $\pm$ 9.1k & -- \\
MMR & 366.0 $\pm$ 5.2k & 2.2 $\pm$ 0.1k & 11.8 $\pm$ 0.1k & -- & 250.8 $\pm$ 4.5k & -- \\

\midrule
\multicolumn{7}{l}{\textbf{Two-stage methods: Post-Retrieval + Pre-Synthesis Pruning}} \\
Centroid Drift & 118.1 $\pm$ 1.3k & 2.3 $\pm$ 0.1k & 2.8 $\pm$ 0.0k & -- & 109.5 $\pm$ 1.2k & 3.6 $\pm$ 0.0k \\
Submodular Coverage & 142.1 $\pm$ 3.2k & 2.2 $\pm$ 0.1k & 3.6 $\pm$ 0.1k & -- & 130.7 $\pm$ 2.4k & 5.5 $\pm$ 0.2k \\
CD + SC & 137.5 $\pm$ 2.0k & 2.3 $\pm$ 0.1k & 3.4 $\pm$ 0.1k & -- & 127.6 $\pm$ 1.6k & 4.2 $\pm$ 0.1k \\
LLM & 226.6 $\pm$ 8.1k & 2.3 $\pm$ 0.1k & 5.8 $\pm$ 0.3k & 25.7 $\pm$ 0.9k & 192.8 $\pm$ 5.4k & -- \\
Lexical & 115.1 $\pm$ 1.2k & 2.3 $\pm$ 0.1k & 2.8 $\pm$ 0.0k & -- & 110.1 $\pm$ 1.3k & -- \\
SC+LLM & 139.1 $\pm$ 4.1k & 2.2 $\pm$ 0.1k & 3.5 $\pm$ 0.1k & 2.8 $\pm$ 0.2k & 88.1 $\pm$ 2.7k & 5.3 $\pm$ 0.2k \\
Hybrid & 121.7 $\pm$ 2.8k & 2.2 $\pm$ 0.1k & 2.9 $\pm$ 0.1k & -- & 77.7 $\pm$ 1.9k & 5.9 $\pm$ 0.2k \\
Combined & 117.6 $\pm$ 2.6k & 2.2 $\pm$ 0.1k & 2.8 $\pm$ 0.1k & -- & 74.9 $\pm$ 1.8k & 5.7 $\pm$ 0.1k \\
CD+LLM & 136.0 $\pm$ 3.2k & 2.2 $\pm$ 0.1k & 3.4 $\pm$ 0.1k & 3.2 $\pm$ 0.1k & 86.6 $\pm$ 2.1k & 4.1 $\pm$ 0.1k \\
MMR & 114.6 $\pm$ 2.2k & 2.2 $\pm$ 0.1k & 2.7 $\pm$ 0.1k & -- & 74.0 $\pm$ 1.6k & 4.3 $\pm$ 0.1k \\

\midrule
\multicolumn{7}{l}{\textbf{Three-stage methods: Pre-Retrieval + Post-Retrieval + Pre-Synthesis Pruning}} \\
Geo.\ Residual Novelty & 150.0 $\pm$ 4.5k & 2.2 $\pm$ 0.1k & 4.4 $\pm$ 0.2k & -- & 138.6 $\pm$ 3.3k & 4.7 $\pm$ 0.1k \\
DPP  & 113.3 $\pm$ 2.2k & 2.3 $\pm$ 0.1k & 3.1 $\pm$ 0.1k & -- & 103.5 $\pm$ 1.6k & 4.4 $\pm$ 0.1k \\
Centroid Drift & 120.6 $\pm$ 2.1k & 2.3 $\pm$ 0.1k & 3.4 $\pm$ 0.1k & -- & 111.2 $\pm$ 1.6k & 3.7 $\pm$ 0.1k \\
Lexical + CD + SC & 120.1 $\pm$ 2.3k & 2.3 $\pm$ 0.1k & 3.5 $\pm$ 0.1k & -- & 110.7 $\pm$ 1.7k & 3.7 $\pm$ 0.1k \\
Lexical & 99.0 $\pm$ 1.1k & 2.3 $\pm$ 0.1k & 2.8 $\pm$ 0.0k & -- & 94.0 $\pm$ 1.1k & -- \\
LLM & 143.9 $\pm$ 7.7k & 2.0 $\pm$ 0.1k & 3.8 $\pm$ 0.2k & 6.2 $\pm$ 0.3k & 82.5 $\pm$ 4.6k & -- \\
Combined & 102.8 $\pm$ 2.2k & 2.2 $\pm$ 0.1k & 2.8 $\pm$ 0.1k & -- & 65.1 $\pm$ 1.5k & 5.8 $\pm$ 0.1k \\
Hybrid & 106.3 $\pm$ 2.6k & 2.2 $\pm$ 0.1k & 2.9 $\pm$ 0.1k & -- & 67.0 $\pm$ 1.8k & 6.0 $\pm$ 0.1k \\
MMR & 100.1 $\pm$ 2.0k & 2.2 $\pm$ 0.1k & 2.7 $\pm$ 0.0k & -- & 64.5 $\pm$ 1.4k & 3.8 $\pm$ 0.1k \\

\bottomrule
\end{tabular}
}
\end{table*}

\begin{table*}[t]
\centering
\caption{Node-level pruning summary}
\label{tab:appendix_node_summary}
\small
\setlength{\tabcolsep}{5pt}
\renewcommand{\arraystretch}{1.15}

\begin{tabular}{lccc}
\toprule
\textbf{Method} & \textbf{\# Nodes} (\S~\ref{sec:num_nodes}) & \textbf{Avg. Pruned Nodes} & \textbf{Pruning Rate (\%)} \\
\midrule

Baseline & 29.0 ± 0.0 & 0.0 ± 0.0 & 0.0 \\
\midrule
\multicolumn{4}{l}{\textbf{One-stage methods: Post-Retrieval Pruning}} \\
Geo.\ Residual Novelty & 13.31 $\pm$ 0.37 & 4.48 $\pm$ 0.10 & 33.7 \\
Centroid Drift & 10.47 $\pm$ 0.13 & 3.44 $\pm$ 0.05 & 32.9 \\
Submodular Coverage & 10.82 $\pm$ 0.28 & 4.82 $\pm$ 0.07 & 44.5 \\
DPP & 9.88 $\pm$ 0.16 & 4.99 $\pm$ 0.06 & 50.5 \\
LLM & 14.90 $\pm$ 0.53 & 6.30 $\pm$ 0.19 & 42.3 \\
Hybrid & 9.28 $\pm$ 0.15 & 4.82 $\pm$ 0.05 & 51.9 \\
Combined & 9.00 $\pm$ 0.11 & 4.75 $\pm$ 0.05 & 52.8 \\
MMR & 8.84 $\pm$ 0.08 & 4.93 $\pm$ 0.03 & 55.8 \\

\midrule
\multicolumn{4}{l}{\textbf{One-stage methods: Pre-Synthesis Pruning
}} \\
Geo.\ Residual Novelty & 28.88 $\pm$ 0.12 & 0.00 $\pm$ 0.00 & 0.0 \\
Centroid Drift & 28.94 $\pm$ 0.05 & 0.00 $\pm$ 0.00 & 0.0 \\
Submodular Coverage & 28.85 $\pm$ 0.13 & 0.00 $\pm$ 0.00 & 0.0 \\
DPP & 28.90 $\pm$ 0.10 & 0.00 $\pm$ 0.00 & 0.0 \\
LLM & 28.98 $\pm$ 0.02 & 0.00 $\pm$ 0.00 & 0.0 \\
Combined & 28.90 $\pm$ 0.08 & 0.00 $\pm$ 0.00 & 0.0 \\
Hybrid & 28.98 $\pm$ 0.02 & 0.00 $\pm$ 0.00 & 0.0 \\
MMR & 28.98 $\pm$ 0.02 & 0.00 $\pm$ 0.00 & 0.0 \\

\midrule
\multicolumn{4}{l}{\textbf{Two-stage methods: Post-Retrieval + Pre-Synthesis Pruning}} \\
Centroid Drift  & 9.02 $\pm$ 0.02 & 4.20 $\pm$ 0.05 & 46.6 \\
Submodular Coverage & 10.80 $\pm$ 0.28 & 4.80 $\pm$ 0.07 & 44.4 \\
CD + SC & 10.45 $\pm$ 0.13 & 3.43 $\pm$ 0.05 & 32.8 \\
LLM & 15.57 $\pm$ 0.53 & 6.30 $\pm$ 0.16 & 40.5 \\
Lexical & 9.00 $\pm$ 0.00 & 4.99 $\pm$ 0.01 & 55.4 \\
SC + LLM & 10.54 $\pm$ 0.30 & 4.79 $\pm$ 0.07 & 45.4 \\
CD + LLM & 10.24 $\pm$ 0.17 & 3.45 $\pm$ 0.05 & 33.7 \\
Combined & 8.98 $\pm$ 0.11 & 4.76 $\pm$ 0.05 & 53.0 \\
Hybrid & 9.30 $\pm$ 0.15 & 4.82 $\pm$ 0.05 & 51.8 \\
MMR & 8.84 $\pm$ 0.08 & 4.94 $\pm$ 0.03 & 55.9 \\

\midrule
\multicolumn{4}{l}{\textbf{Three-stage methods: Pre-Retrieval + Post-Retrieval + Pre-Synthesis Pruning}} \\
Geo.\ Residual Novelty & 11.57 $\pm$ 0.33 & 3.49 $\pm$ 0.09 & 30.2 \\
DPP & 8.68 $\pm$ 0.15 & 3.95 $\pm$ 0.05 & 45.5 \\
Centroid Drift & 9.27 $\pm$ 0.13 & 2.45 $\pm$ 0.05 & 26.4 \\
Lexical + CD + SC & 9.26 $\pm$ 0.14 & 2.31 $\pm$ 0.06 & 24.9 \\
Lexical & 7.90 $\pm$ 0.04 & 3.88 $\pm$ 0.04 & 49.1 \\
LLM & 10.12 $\pm$ 0.45 & 4.57 $\pm$ 0.13 & 45.2 \\
Combined & 7.90 $\pm$ 0.10 & 3.75 $\pm$ 0.05 & 47.5 \\
Hybrid & 8.16 $\pm$ 0.14 & 3.81 $\pm$ 0.05 & 46.7 \\
MMR & 7.82 $\pm$ 0.08 & 3.97 $\pm$ 0.02 & 50.8 \\

\bottomrule
\end{tabular}

\vspace{4pt}

\begin{minipage}{0.95\textwidth}
\footnotesize
This table summarizes how strongly each method contracts the research tree. \# Nodes estimates the number of nodes actually explored per report, while Avg.\ Pruned Nodes measures how many of those explored nodes were discarded. Pruning Rate therefore captures the fraction of explored nodes removed by the pruning policy.
\end{minipage}

\end{table*}

\begin{table*}[t]
\centering
\caption{Token accounting summary (k tokens; mean $\pm$ SE). Mean Savings vs. Baseline is computed from the reported mean total token counts.}
\label{tab:appendix_token_accounting}
\small
\setlength{\tabcolsep}{5pt}
\renewcommand{\arraystretch}{1.15}

\begin{tabular}{lccccc}
\toprule
\textbf{Method} & \textbf{Input} & \textbf{Output} & \textbf{Total} & \textbf{Est. Saved} & \textbf{Savings vs. Baseline (k)} \\
\midrule

Baseline & 260.7 $\pm$ 2.7 & 114.7 $\pm$ 0.7 & 375.4 $\pm$ 2.6 & 149.1 $\pm$ 1.8 & 0.0 \\

\midrule
\multicolumn{6}{l}{\textbf{One-stage methods: Post-Retrieval Pruning}} \\
Geo.\ Residual Novelty & 125.3 $\pm$ 3.7 & 50.4 $\pm$ 1.6 & 175.7 $\pm$ 5.1 & 51.0 $\pm$ 2.1 & 199.7 \\
Centroid Drift & 99.0 $\pm$ 1.6 & 38.5 $\pm$ 0.6 & 137.5 $\pm$ 2.0 & 32.4 $\pm$ 0.8 & 237.9 \\
Submodular Coverage & 101.8 $\pm$ 2.1 & 40.1 $\pm$ 1.3 & 141.9 $\pm$ 3.3 & 38.7 $\pm$ 0.9 & 233.4 \\
DPP & 93.7 $\pm$ 1.8 & 36.0 $\pm$ 0.8 & 129.6 $\pm$ 2.5 & 36.4 $\pm$ 0.7 & 245.7 \\
LLM & 154.0 $\pm$ 5.7 & 57.8 $\pm$ 2.3 & 211.8 $\pm$ 7.9 & 63.4 $\pm$ 3.0 & 163.5 \\
Hybrid & 87.7 $\pm$ 2.2 & 33.5 $\pm$ 0.7 & 121.3 $\pm$ 2.8 & 33.5 $\pm$ 0.9 & 67.7 \\
MMR & 82.9 $\pm$ 1.9 & 31.7 $\pm$ 0.5 & 114.6 $\pm$ 2.2 & 34.2 $\pm$ 0.8 & 69.5 \\
Combined & 85.5 $\pm$ 2.1 & 32.3 $\pm$ 0.6 & 117.8 $\pm$ 2.5 & 33.0 $\pm$ 0.9 & 257.6 \\

\midrule
\multicolumn{6}{l}{\textbf{One-stage methods: Pre-Synthesis Pruning}} \\
Geo.\ Residual Novelty & 259.5 $\pm$ 2.8 & 114.5 $\pm$ 0.9 & 374.1 $\pm$ 3.1 & 147.1 $\pm$ 2.0 & 1.3 \\
Centroid Drift & 259.7 $\pm$ 2.7 & 114.8 $\pm$ 0.7 & 374.5 $\pm$ 2.8 & 147.2 $\pm$ 1.9 & 0.9 \\
Submodular Coverage & 270.2 $\pm$ 2.9 & 114.4 $\pm$ 0.9 & 384.7 $\pm$ 3.3 & 147.3 $\pm$ 2.0 & -9.3 \\
DPP & 259.8 $\pm$ 3.0 & 114.6 $\pm$ 0.8 & 374.4 $\pm$ 3.2 & 147.3 $\pm$ 1.9 & 1.0 \\
LLM & 271.3 $\pm$ 3.1 & 115.5 $\pm$ 0.7 & 386.7 $\pm$ 3.1 & 149.2 $\pm$ 2.2 & -11.3 \\
Combined & 260.4 $\pm$ 5.7 & 114.1 $\pm$ 0.9 & 374.5 $\pm$ 6.1 & 171.6 $\pm$ 4.2 & 0.2 \\
Hybrid & 221.9 $\pm$ 9.9 & 110.4 $\pm$ 1.0 & 332.3 $\pm$ 10.6 & 137.4 $\pm$ 7.3 & 11.5 \\
MMR & 251.5 $\pm$ 5.0 & 114.5 $\pm$ 0.8 & 366.0 $\pm$ 5.2 & 142.3 $\pm$ 3.4 & 2.5 \\

\midrule
\multicolumn{6}{l}{\textbf{Two-stage methods: Post-Retrieval + Pre-Synthesis Pruning}} \\
Centroid Drift & 85.9 $\pm$ 1.3 & 32.2 $\pm$ 0.3 & 118.1 $\pm$ 1.3 & 30.8 $\pm$ 0.5 & 257.3 \\
Submodular Coverage & 102.4 $\pm$ 2.2 & 39.8 $\pm$ 1.3 & 142.1 $\pm$ 3.2 & 41.9 $\pm$ 0.9 & 233.3 \\
CD + SC & 99.1 $\pm$ 1.6 & 38.4 $\pm$ 0.6 & 137.5 $\pm$ 2.0 & 36.1 $\pm$ 0.9 & 237.9 \\
LLM & 165.6 $\pm$ 5.8 & 60.9 $\pm$ 2.3 & 226.6 $\pm$ 8.1 & 70.0 $\pm$ 2.9 & 148.8 \\
Lexical & 82.8 $\pm$ 1.3 & 32.4 $\pm$ 0.3 & 115.1 $\pm$ 1.2 & 48.0 $\pm$ 0.7 & 260.3 \\
SC + LLM & 99.9 $\pm$ 2.9 & 39.2 $\pm$ 1.4 & 139.1 $\pm$ 4.1 & 41.7 $\pm$ 1.3 & 236.3 \\
Combined & 85.3 $\pm$ 2.1 & 32.3 $\pm$ 0.6 & 117.6 $\pm$ 2.6 & 47.9 $\pm$ 1.3 & 68.7 \\
CD + LLM & 98.1 $\pm$ 2.5 & 37.9 $\pm$ 0.8 & 136.0 $\pm$ 3.2 & 36.5 $\pm$ 1.2 & 239.4 \\
Hybrid & 88.0 $\pm$ 2.2 & 33.6 $\pm$ 0.7 & 121.7 $\pm$ 2.8 & 46.1 $\pm$ 1.5 & 67.6 \\
MMR & 82.9 $\pm$ 1.9 & 31.7 $\pm$ 0.5 & 114.6 $\pm$ 2.2 & 34.2 $\pm$ 0.8 & 69.5 \\

\midrule
\multicolumn{6}{l}{\textbf{Three-stage methods: Pre-Retrieval + Post-Retrieval + Pre-Synthesis Pruning}} \\
Geo.\ Residual Novelty & 106.3 $\pm$ 3.2 & 43.7 $\pm$ 1.4 & 150.0 $\pm$ 4.5 & 38.7 $\pm$ 1.7 & 225.4 \\
DPP & 81.6 $\pm$ 1.7 & 31.7 $\pm$ 0.7 & 113.3 $\pm$ 2.2 & 28.5 $\pm$ 0.6 & 262.0 \\
Centroid Drift & 86.4 $\pm$ 1.7 & 34.2 $\pm$ 0.5 & 120.6 $\pm$ 2.1 & 24.4 $\pm$ 0.7 & 254.8 \\
Lexical + CD + SC & 86.0 $\pm$ 1.7 & 34.1 $\pm$ 0.6 & 120.1 $\pm$ 2.3 & 26.5 $\pm$ 0.9 & 255.3 \\
Lexical & 70.7 $\pm$ 1.0 & 28.3 $\pm$ 0.3 & 99.0 $\pm$ 1.1 & 40.2 $\pm$ 0.7 & 276.4 \\
LLM & 105.0 $\pm$ 5.8 & 38.9 $\pm$ 2.0 & 143.9 $\pm$ 7.7 & 38.9 $\pm$ 2.2 & 231.5 \\
Combined & 74.3 $\pm$ 1.8 & 28.5 $\pm$ 0.5 & 102.8 $\pm$ 2.2 & 40.9 $\pm$ 1.1 & 72.6 \\
Hybrid & 76.8 $\pm$ 2.1 & 29.5 $\pm$ 0.6 & 106.3 $\pm$ 2.6 & 38.9 $\pm$ 1.4 & 71.7 \\
MMR & 72.0 $\pm$ 1.7 & 28.2 $\pm$ 0.5 & 100.1 $\pm$ 2.0 & 27.0 $\pm$ 0.7 & 73.3 \\

\bottomrule
\end{tabular}

\vspace{4pt}

\begin{minipage}{0.95\textwidth}
\footnotesize
This table provides a more detailed view of token usage. In addition to total token cost, it separates input and output tokens, reports the estimated number of tokens saved by pruning, and shows the absolute mean token savings relative to the baseline. This helps distinguish methods that achieve low total cost by aggressively pruning large prompt contexts from those that reduce generation cost or avoid expanding large parts of the search tree altogether.
\end{minipage}

\end{table*}

\begin{table*}[t]
\centering
\caption{Share of total token budget by pipeline stage (\%; mean ± SE). Dashes indicate stages that do not consume logged tokens for that method.}
\label{tab:appendix_token_shares_10}
\small
\setlength{\tabcolsep}{5pt}
\renewcommand{\arraystretch}{1.15}

\begin{tabular}{lccccc}
\toprule
\textbf{Method} & \textbf{Planning} & \textbf{Query Generation} & \textbf{Query Pruning} & \textbf{Result Proc.} & \textbf{Embedding} \\
\midrule

Baseline & 1.02 ± 0.05 & 4.58 ± 0.14 & -- & \textbf{94.41 ± 0.18} & -- \\
\midrule
\multicolumn{6}{l}{\textbf{One-stage methods: Post-Retrieval Pruning}} \\
Geo.\ Residual Novelty & 1.42 $\pm$ 0.07 & 2.68 $\pm$ 0.04 & -- & 92.76 $\pm$ 0.40 & 3.14 $\pm$ 0.03 \\
Centroid Drift & 1.69 $\pm$ 0.07 & 2.54 $\pm$ 0.04 & -- & 92.68 $\pm$ 0.44 & 3.10 $\pm$ 0.03 \\
Submodular Coverage & 1.67 $\pm$ 0.07 & 2.50 $\pm$ 0.05 & -- & 91.98 $\pm$ 0.45 & 3.85 $\pm$ 0.04 \\
DPP & 1.81 $\pm$ 0.07 & 2.47 $\pm$ 0.04 & -- & 91.90 $\pm$ 0.42 & 3.82 $\pm$ 0.04 \\
LLM & 1.23 $\pm$ 0.06 & 2.52 $\pm$ 0.05 & 9.55 $\pm$ 0.14 & 86.70 $\pm$ 0.44 & -- \\
Combined & 1.90 $\pm$ 0.07 & 2.50 $\pm$ 0.08 & -- & 62.51 $\pm$ 0.77 & 4.68 $\pm$ 0.11 \\
Hybrid & 1.86 $\pm$ 0.07 & 2.55 $\pm$ 0.08 & -- & 62.84 $\pm$ 0.67 & 4.70 $\pm$ 0.11 \\
MMR & 1.93 $\pm$ 0.06 & 2.53 $\pm$ 0.09 & -- & 63.42 $\pm$ 0.72 & 3.63 $\pm$ 0.08 \\

\midrule
\multicolumn{6}{l}{\textbf{One-stage methods: Pre-Synthesis Pruning}} \\
Geo.\ Residual Novelty & 0.60 $\pm$ 0.02 & 3.19 $\pm$ 0.04 & -- & 96.21 $\pm$ 0.39 & -- \\
Centroid Drift & 0.60 $\pm$ 0.02 & 3.19 $\pm$ 0.04 & -- & 96.21 $\pm$ 0.40 & -- \\
Submodular Coverage & 0.60 $\pm$ 0.02 & 3.10 $\pm$ 0.04 & -- & 93.22 $\pm$ 0.39 & 3.09 $\pm$ 0.03 \\
DPP & 0.61 $\pm$ 0.02 & 3.19 $\pm$ 0.04 & -- & 96.21 $\pm$ 0.48 & -- \\
LLM & 0.60 $\pm$ 0.02 & 3.10 $\pm$ 0.04 & 3.43 $\pm$ 0.06 & 92.86 $\pm$ 0.45 & -- \\
Combined & 0.58 $\pm$ 0.02 & 3.31 $\pm$ 0.11 & -- & 65.21 $\pm$ 0.83 & -- \\
Hybrid & 0.54 $\pm$ 0.02 & 4.10 $\pm$ 0.20 & -- & 58.85 $\pm$ 1.56 & -- \\
MMR & 0.60 $\pm$ 0.02 & 3.36 $\pm$ 0.09 & -- & 67.70 $\pm$ 0.76 & -- \\

\midrule
\multicolumn{6}{l}{\textbf{Two-stage methods: Post-Retrieval + Pre-Synthesis Pruning}} \\
Centroid Drift & 1.93 $\pm$ 0.07 & 2.40 $\pm$ 0.04 & -- & 92.61 $\pm$ 0.59 & 3.06 $\pm$ 0.03 \\
Submodular Coverage & 1.67 $\pm$ 0.07 & 2.49 $\pm$ 0.05 & -- & 91.99 $\pm$ 0.58 & 3.84 $\pm$ 0.04 \\
CD + SC & 1.68 $\pm$ 0.07 & 2.53 $\pm$ 0.04 & -- & 92.69 $\pm$ 0.45 & 3.10 $\pm$ 0.03 \\
LLM & 1.16 $\pm$ 0.06 & 2.48 $\pm$ 0.04 & 11.13 $\pm$ 0.18 & 85.25 $\pm$ 0.48 & -- \\
Lexical & 1.98 $\pm$ 0.07 & 2.45 $\pm$ 0.05 & -- & 95.57 $\pm$ 0.09 & -- \\
SC+LLM & 1.68 $\pm$ 0.07 & 2.57 $\pm$ 0.07 & 1.87 $\pm$ 0.06 & 62.39 $\pm$ 0.69 & 3.61 $\pm$ 0.08 \\
CD+LLM & 1.67 $\pm$ 0.06 & 2.60 $\pm$ 0.08 & 2.29 $\pm$ 0.06 & 62.57 $\pm$ 0.70 & 2.90 $\pm$ 0.07 \\
Hybrid & 1.84 $\pm$ 0.06 & 2.53 $\pm$ 0.08 & -- & 62.69 $\pm$ 0.74 & 4.73 $\pm$ 0.11 \\
Combined & 1.90 $\pm$ 0.07 & 2.52 $\pm$ 0.09 & -- & 62.40 $\pm$ 0.82 & 4.67 $\pm$ 0.11 \\
MMR & 1.93 $\pm$ 0.06 & 2.51 $\pm$ 0.08 & -- & 63.43 $\pm$ 0.72 & 3.63 $\pm$ 0.08 \\

\midrule
\multicolumn{6}{l}{\textbf{Three-stage methods: Pre-Retrieval + Post-Retrieval + Pre-Synthesis Pruning}} \\
Geo.\ Residual Novelty & 1.62 $\pm$ 0.07 & 2.95 $\pm$ 0.04 & -- & 92.28 $\pm$ 0.44 & 3.15 $\pm$ 0.03 \\
DPP & 2.07 $\pm$ 0.08 & 2.78 $\pm$ 0.04 & -- & 91.25 $\pm$ 0.46 & 3.90 $\pm$ 0.04 \\
Centroid Drift & 1.94 $\pm$ 0.08 & 2.85 $\pm$ 0.04 & -- & 92.12 $\pm$ 0.52 & 3.09 $\pm$ 0.03 \\
Lexical + CD + SC & 1.94 $\pm$ 0.08 & 2.92 $\pm$ 0.06 & -- & 92.04 $\pm$ 0.45 & 3.10 $\pm$ 0.03 \\
Lexical & 2.30 $\pm$ 0.08 & 2.88 $\pm$ 0.06 & -- & 94.82 $\pm$ 0.11 & -- \\
LLM & 1.74 $\pm$ 0.10 & 3.04 $\pm$ 0.15 & 4.99 $\pm$ 0.25 & 53.72 $\pm$ 1.17 & -- \\
Combined & 2.18 $\pm$ 0.08 & 2.91 $\pm$ 0.12 & -- & 62.00 $\pm$ 0.78 & 5.75 $\pm$ 0.07 \\
Hybrid & 2.13 $\pm$ 0.08 & 2.91 $\pm$ 0.10 & -- & 61.71 $\pm$ 0.79 & 5.72 $\pm$ 0.06 \\
MMR & 2.24 $\pm$ 0.08 & 2.96 $\pm$ 0.12 & -- & 63.30 $\pm$ 0.72 & 3.68 $\pm$ 0.08 \\

\bottomrule
\end{tabular}

\vspace{4pt}

\begin{minipage}{0.95\textwidth}
\footnotesize
This table decomposes each method's total token budget into stage-wise shares, highlighting where token usage is concentrated within the pipeline. While most methods are dominated by result-processing costs, approaches with explicit LLM-based pre-retrieval pruning allocate a significant fraction of tokens to the pruning stage itself.
\end{minipage}

\end{table*}

\begin{table*}[t]
\centering
\caption{Pruning-stage effectiveness (\%; mean ± SE). Ratios denote the fraction of candidate items removed at each stage; token reduction denotes the corresponding decrease in context tokens.}
\label{tab:appendix_pruning_effectiveness}
\small
\setlength{\tabcolsep}{4pt}
\renewcommand{\arraystretch}{1.15}
\resizebox{\textwidth}{!}{%
\begin{tabular}{lcccccc}
\toprule
\textbf{Method} & \textbf{Q-Ratio} & \textbf{Q-Token Red.} & \textbf{Branch Ratio} & \textbf{Branch Token Red.} & \textbf{Root Ratio} & \textbf{Root Token Red.} \\
\midrule

Baseline & -- & -- & -- & -- & -- & -- \\
\midrule
\multicolumn{7}{l}{\textbf{One-stage methods: Post-Retrieval Pruning}} \\
Geo.\ Residual Novelty & -- & -- & 39.6 $\pm$ 1.4 & 40.6 $\pm$ 1.4 & 11.3 $\pm$ 1.0 & 12.1 $\pm$ 1.1 \\
Centroid Drift & -- & -- & 37.5 $\pm$ 1.0 & 40.2 $\pm$ 1.1 & 4.7 $\pm$ 0.5 & 5.5 $\pm$ 0.6 \\
Submodular Coverage & -- & -- & 51.7 $\pm$ 1.1 & 54.1 $\pm$ 1.1 & 2.3 $\pm$ 0.5 & 2.5 $\pm$ 0.6 \\
DPP & -- & -- & 57.5 $\pm$ 0.7 & 58.3 $\pm$ 0.8 & 0.8 $\pm$ 0.3 & 1.0 $\pm$ 0.3 \\
LLM & -- & -- & 50.7 $\pm$ 1.9 & 49.7 $\pm$ 1.9 & 12.2 $\pm$ 1.2 & 12.9 $\pm$ 1.2 \\
Combined & -- & -- & 60.5 $\pm$ 1.0 & 61.4 $\pm$ 1.0 & 0.2 $\pm$ 0.1 & 0.2 $\pm$ 0.2 \\
Hybrid & -- & -- & 59.9 $\pm$ 1.1 & 60.6 $\pm$ 1.1 & 0.4 $\pm$ 0.2 & 0.6 $\pm$ 0.2 \\
MMR & -- & -- & 63.9 $\pm$ 0.8 & 64.6 $\pm$ 0.8 & 0.0 $\pm$ 0.0 & 0.0 $\pm$ 0.0 \\
\midrule
\multicolumn{7}{l}{\textbf{One-stage methods: Pre-Synthesis Pruning}} \\
Geo.\ Residual Novelty & -- & -- & -- & -- & 75.8 $\pm$ 0.4 & 77.1 $\pm$ 0.3 \\
Centroid Drift & -- & -- & -- & -- & 75.9 $\pm$ 0.3 & 77.2 $\pm$ 0.2 \\
Submodular Coverage & -- & -- & -- & -- & 75.3 $\pm$ 0.4 & 77.5 $\pm$ 0.3 \\
DPP & -- & -- & -- & -- & 76.2 $\pm$ 0.4 & 77.5 $\pm$ 0.2 \\
LLM & -- & -- & -- & -- & 62.9 $\pm$ 0.5 & 61.8 $\pm$ 0.5 \\
Combined & -- & -- & -- & -- & 96.1 $\pm$ 0.1 & 96.0 $\pm$ 0.1 \\
Hybrid & -- & -- & -- & -- & 92.6 $\pm$ 0.8 & 92.5 $\pm$ 0.8 \\
MMR & -- & -- & -- & -- & 77.2 $\pm$ 0.5 & 76.2 $\pm$ 0.5 \\
\midrule
\multicolumn{7}{l}{\textbf{Two-stage methods: Post-Retrieval + Pre-Synthesis Pruning}} \\
Centroid Drift & -- & -- & 52.6 $\pm$ 0.6 & 54.4 $\pm$ 0.7 & 0.0 $\pm$ 0.0 & 0.0 $\pm$ 0.0 \\
Submodular Coverage & -- & -- & 51.6 $\pm$ 1.1 & 53.9 $\pm$ 1.1 & 16.3 $\pm$ 1.7 & 17.0 $\pm$ 1.9 \\
CD + SC & -- & -- & 37.6 $\pm$ 1.0 & 40.2 $\pm$ 1.0 & 20.3 $\pm$ 1.7 & 21.0 $\pm$ 1.7 \\
LLM & -- & -- & 48.6 $\pm$ 1.9 & 48.0 $\pm$ 1.9 & 26.1 $\pm$ 1.7 & 24.7 $\pm$ 1.8 \\
Lexical & -- & -- & 58.2 $\pm$ 0.7 & 57.7 $\pm$ 0.9 & 60.0 $\pm$ 1.6 & 57.9 $\pm$ 1.7 \\
MMR & -- & -- & 63.9 $\pm$ 0.8 & 64.6 $\pm$ 0.8 & 0.0 $\pm$ 0.0 & 0.0 $\pm$ 0.0 \\
Hybrid & -- & -- & 59.6 $\pm$ 1.1 & 60.4 $\pm$ 1.1 & 54.4 $\pm$ 3.1 & 53.8 $\pm$ 3.1 \\
Combined & -- & -- & 60.9 $\pm$ 1.0 & 61.6 $\pm$ 1.0 & 68.6 $\pm$ 0.9 & 67.1 $\pm$ 1.1 \\
SC+LLM & -- & -- & 53.8 $\pm$ 1.4 & 56.0 $\pm$ 1.4 & 23.7 $\pm$ 1.8 & 22.3 $\pm$ 1.7 \\
CD+LLM & -- & -- & 40.0 $\pm$ 1.6 & 42.5 $\pm$ 1.6 & 25.9 $\pm$ 0.6 & 20.3 $\pm$ 0.5 \\

\midrule
\multicolumn{7}{l}{\textbf{Three-stage methods: Pre-Retrieval + Post-Retrieval + Pre-Synthesis Pruning}} \\
Geo.\ Residual Novelty & 25.0 $\pm$ 0.0 & 24.7 $\pm$ 0.3 & 36.0 $\pm$ 1.3 & 36.2 $\pm$ 1.3 & 20.4 $\pm$ 1.9 & 21.2 $\pm$ 2.0 \\
DPP & 25.0 $\pm$ 0.0 & 24.2 $\pm$ 0.2 & 52.3 $\pm$ 0.7 & 52.4 $\pm$ 0.8 & 1.4 $\pm$ 0.6 & 1.4 $\pm$ 0.6 \\
Centroid Drift & 25.0 $\pm$ 0.0 & 24.8 $\pm$ 0.3 & 31.1 $\pm$ 1.1 & 32.5 $\pm$ 1.1 & 9.5 $\pm$ 1.0 & 9.6 $\pm$ 1.1 \\
Lexical + CD + SC & 27.8 $\pm$ 1.0 & 26.9 $\pm$ 1.1 & 29.1 $\pm$ 1.0 & 30.8 $\pm$ 1.0 & 17.8 $\pm$ 1.5 & 18.4 $\pm$ 1.5 \\
Lexical & 27.5 $\pm$ 1.0 & 26.6 $\pm$ 1.0 & 54.5 $\pm$ 0.6 & 53.5 $\pm$ 0.8 & 59.0 $\pm$ 1.6 & 58.2 $\pm$ 1.8 \\
MMR & 27.2 $\pm$ 0.5 & 21.2 $\pm$ 0.4 & 59.2 $\pm$ 0.9 & 58.8 $\pm$ 0.9& 0.0 $\pm$ 0.0 & 0.0 $\pm$ 0.0 \\
LLM & 27.0 $\pm$ 0.9 & 21.3 $\pm$ 0.8 & 60.5 $\pm$ 2.5 & 59.2 $\pm$ 2.5 & 22.5 $\pm$ 2.3 & 21.1 $\pm$ 2.3 \\
Hybrid & 33.6 $\pm$ 0.8 & 26.8 $\pm$ 0.7 & 25.0 $\pm$ 0.0 & 24.0 $\pm$ 0.2 & 55.1 $\pm$ 1.2 & 55.0 $\pm$ 1.2 \\
Combined & 35.5 $\pm$ 0.7 & 28.5 $\pm$ 0.6 & 25.0 $\pm$ 0.0 & 24.0 $\pm$ 0.2 & 55.8 $\pm$ 1.1 & 55.7 $\pm$ 1.2 \\

\bottomrule
\end{tabular}%
}
\vspace{4pt}

\begin{minipage}{0.95\textwidth}
\footnotesize
This table decomposes pruning behavior across pre-retrieval, post-retrieval, and pre-synthesis stages, reporting both the fraction of items removed and the corresponding reduction in token load. This breakdown reveals where pruning occurs within the pipeline, distinguishing methods that act early at the pre-retrieval stage from those relying primarily on post-retrieval or pre-synthesis reduction.
\end{minipage}

\end{table*}

\begin{figure*}[t]
    \centering
    \includegraphics[width=0.9\textwidth]{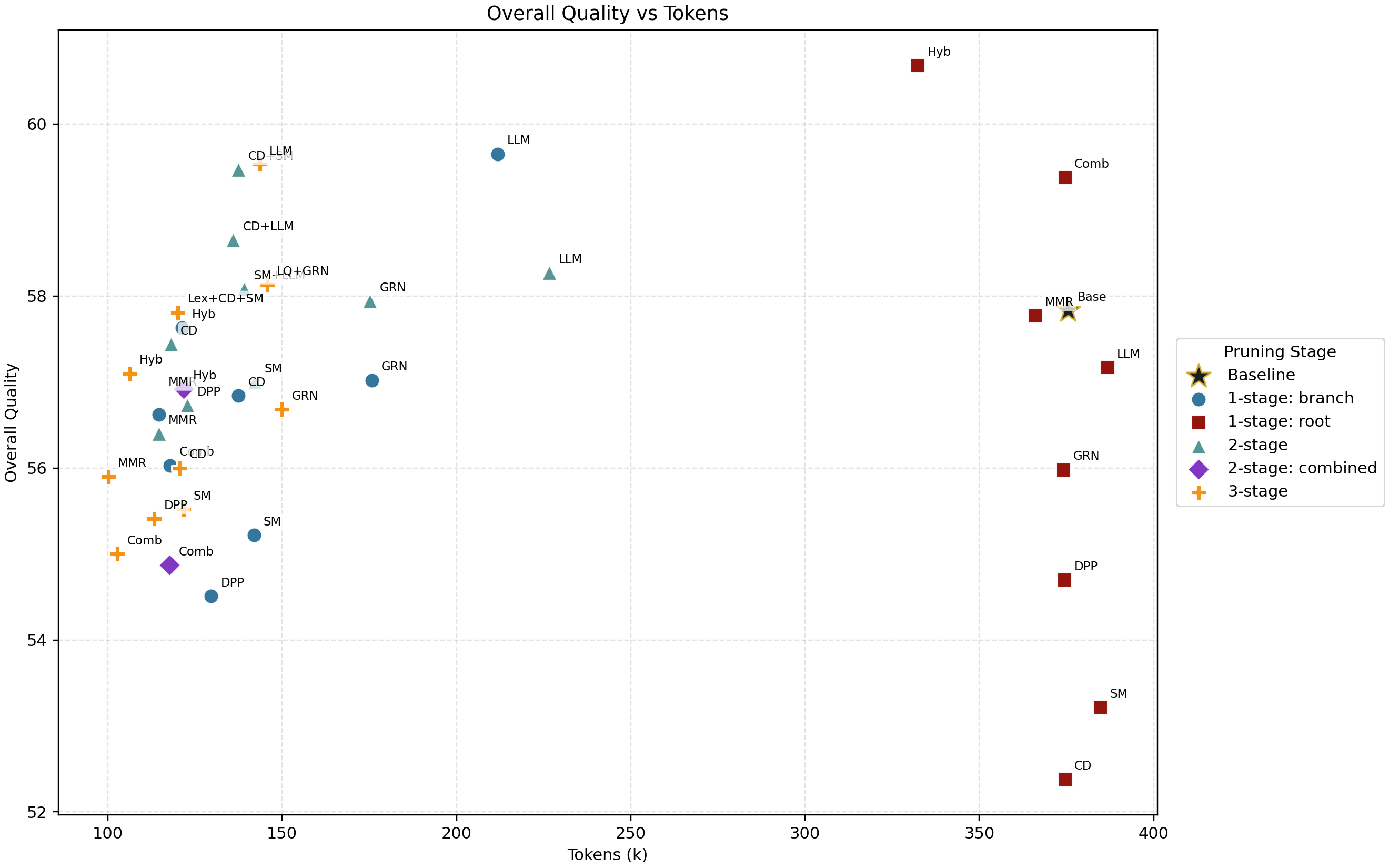}
    \caption{Overall quality versus token usage in thousands (averaged over the 100 reports) for all pruning strategies reported in Table~\ref{tab:quality_all}. Each point corresponds to one method configuration, with marker style and color indicating the pruning stage. The baseline is highlighted separately, and labels use shortened method names for readability.}
    \label{fig:quality_vs_tokens_all}
\end{figure*}

\begin{table*}[t]
\centering
\caption{DeepResearch Bench overall quality across pruning criteria and stage placement. Scores are RACE-style report-quality metrics. Bold denotes the best pruned value in each stage column.}
\label{tab:bench_quality_summary}
\small
\setlength{\tabcolsep}{8pt}
\renewcommand{\arraystretch}{1.10}
\begin{tabular}{lcccc}
\toprule
\textbf{Criterion} & \textbf{Branch-only} & \textbf{Root-only} & \textbf{Two-stage} & \textbf{Three-stage} \\
\midrule
Baseline & 0.4798 & 0.4798 & 0.4798 & 0.4798 \\
\midrule
DPP & 0.4584 & 0.4774 & 0.4648 & 0.4591 \\
Centroid Drift & 0.4567 & 0.4759 & 0.4676 & 0.4590 \\
GRN & \textbf{0.4688} & 0.4794 & \textbf{0.4716} & \textbf{0.4612} \\
MMR & 0.4541 & 0.4766 & 0.4605 & 0.4523 \\
Submodular & 0.4557 & \textbf{0.4811} & 0.4561 & 0.4534 \\
\bottomrule
\end{tabular}
\end{table*}
\begin{table*}[t]
\centering
\caption{DeepResearch Bench efficiency across pruning criteria and stage placement. Tokens are reported in thousands (k), and savings are relative to the unpruned baseline. Bold denotes the best pruned value in each stage column.}
\label{tab:bench_efficiency_summary}
\scriptsize
\setlength{\tabcolsep}{3pt}
\renewcommand{\arraystretch}{1.08}

\resizebox{\textwidth}{!}{%
\begin{tabular}{lccc ccc ccc ccc}
\toprule
& \multicolumn{3}{c}{\textbf{Branch-only}} 
& \multicolumn{3}{c}{\textbf{Root-only}} 
& \multicolumn{3}{c}{\textbf{Two-stage}} 
& \multicolumn{3}{c}{\textbf{Three-stage}} \\
\cmidrule(lr){2-4} 
\cmidrule(lr){5-7} 
\cmidrule(lr){8-10}
\cmidrule(lr){11-13}
\textbf{Criterion} 
& \textbf{Tokens} & \textbf{Tok. Sav.} & \textbf{Run. Sav.}
& \textbf{Tokens} & \textbf{Tok. Sav.} & \textbf{Run. Sav.}
& \textbf{Tokens} & \textbf{Tok. Sav.} & \textbf{Run. Sav.}
& \textbf{Tokens} & \textbf{Tok. Sav.} & \textbf{Run. Sav.} \\
\midrule
DPP 
& $89.42$ & $72.65\%$ & $81.14\%$ 
& $281.98$ & \textbf{$13.77\%$} & \textbf{$32.54\%$} 
& $89.81$ & $72.53\%$ & $80.95\%$
& $25.60$ & $92.17\%$ & $93.60\%$ \\

Centroid Drift 
& $95.50$ & $70.79\%$ & $78.01\%$ 
& $300.54$ & $8.09\%$ & $28.62\%$ 
& $92.62$ & $71.68\%$ & $80.32\%$
& $81.41$ & $75.10\%$ & $79.93\%$ \\

GRN 
& $110.72$ & $66.14\%$ & $74.77\%$ 
& $286.25$ & $12.46\%$ & $28.75\%$ 
& $90.57$ & $72.30\%$ & \textbf{$82.89\%$}
& $27.27$ & $91.66\%$ & $93.44\%$ \\

MMR 
& \textbf{$86.05$} & \textbf{$73.68\%$} & \textbf{$81.94\%$} 
& $301.83$ & $7.69\%$ & $28.49\%$ 
& \textbf{$89.03$} & \textbf{$72.77\%$} & $81.76\%$
& \textbf{$25.13$} & \textbf{$92.32\%$} & \textbf{$94.30\%$} \\

Submodular 
& $96.62$ & $70.45\%$ & $79.77\%$ 
& $312.15$ & $4.54\%$ & $24.78\%$ 
& $97.68$ & $70.13\%$ & $67.35\%$
& $84.03$ & $74.30\%$ & $82.92\%$ \\
\bottomrule
\end{tabular}%
}

\vspace{0.35em}
\end{table*}

\begin{table*}[t]
\centering
\caption{Local threshold sweeps for representative post-retrieval pruning methods. $\Delta Q_{\mathrm{rel}}$ is the percent change in overall quality relative to the published threshold setting for that method. Rows satisfying the 2\% stability criterion are marked with \checkmark. Tokens are reported in thousands (k). Values are mean scores over a 10-query sensitivity subset.}
\label{tab:threshold_sweep_points}
\scriptsize
\setlength{\tabcolsep}{4pt}
\renewcommand{\arraystretch}{1.06}
\begin{tabular}{l c c c c c c}
\toprule
\textbf{Method} & \textbf{Threshold} & \textbf{\# Tokens} & \textbf{Runtime (s)} & \textbf{Overall} & \textbf{$\Delta Q_{\mathrm{rel}}$} & \textbf{Stable?} \\
\midrule
\multicolumn{7}{l}{\textbf{MMR} (fixed $\lambda = 0.35$, published $\tau_{\mathrm{pub}}=0.35$)} \\
& 0.10 & 121.3 & 1239.0 & 58.33 & +3.24\% & \checkmark \\
& 0.20 & 121.3 & 1238.9 & 57.50 & +1.77\% & \checkmark \\
& 0.30 & 121.3 & 1239.0 & 56.83 & +0.58\% & \checkmark \\
& 0.35 & 121.3 & 1239.1 & 56.50 & 0.00\% & \checkmark \\
& 0.40 & 121.3 & 1239.3 & 56.50 & 0.00\% & \checkmark \\
\midrule
\multicolumn{7}{l}{\textbf{GRN} (published $\tau_{\mathrm{pub}}=0.85$)} \\
& 0.65 & 307.3 & 3034.3 & 61.30 & +0.21\% & \checkmark \\
& 0.75 & 268.7 & 2616.3 & 61.17 & 0.00\% & \checkmark \\
& 0.80 & 231.4 & 2300.3 & 58.17 & -4.90\% &  \\
& 0.85 & 171.9 & 1756.2 & 61.17 & 0.00\% & \checkmark \\
& 0.90 & 138.3 & 1430.1 & 59.83 & -2.19\% &  \\
\midrule
\multicolumn{7}{l}{\textbf{Centroid Drift} (published $\tau_{\mathrm{pub}}=0.03$)} \\
& 0.005 & 227.0 & 2336.6 & 61.50 & +0.28\% & \checkmark \\
& 0.01 & 202.9 & 2068.9 & 60.50 & -1.35\% & \checkmark \\
& 0.03 & 137.4 & 1438.2 & 61.33 & 0.00\% & \checkmark \\
& 0.05 & 126.1 & 1341.5 & 58.67 & -4.34\% &  \\
& 0.085 & 120.4 & 1237.7 & 58.83 & -4.08\% &  \\
\midrule
\multicolumn{7}{l}{\textbf{DPP} (published $\tau_{\mathrm{pub}}=0.30$)} \\
& 0.05 & 391.4 & 3723.4 & 61.33 & +0.82\% & \checkmark \\
& 0.10 & 363.1 & 3465.4 & 60.83 & 0.00\% & \checkmark \\
& 0.20 & 216.0 & 2223.6 & 59.83 & -1.64\% & \checkmark \\
& 0.30 & 134.1 & 1422.9 & 60.83 & 0.00\% & \checkmark \\
& 0.40 & 121.3 & 1237.6 & 57.33 & -5.75\% &  \\
\midrule
\multicolumn{7}{l}{\textbf{Submodular Coverage} (published $\tau_{\mathrm{pub}}=0.05$)} \\
& 0.01 & 383.1 & 3652.0 & 59.83 & -0.83\% & \checkmark \\
& 0.03 & 232.7 & 2303.3 & 59.83 & -0.83\% & \checkmark \\
& 0.05 & 143.7 & 1516.2 & 60.33 & 0.00\% & \checkmark \\
& 0.08 & 126.3 & 1302.8 & 59.17 & -1.92\% & \checkmark \\
& 0.10 & 123.7 & 1261.1 & 57.67 & -4.41\% &  \\
\bottomrule
\end{tabular}
\end{table*}